%% file: main.tex
\documentclass[lettersize,journal]{IEEEtran}
\usepackage{tikz,xcolor,hyperref}
\definecolor{lime}{HTML}{A6CE39}
\DeclareRobustCommand{\orcidicon}{%
    \begin{tikzpicture}
    \draw[lime, fill=lime] (0,0) 
    circle [radius=0.16] 
    node[white] {{\fontfamily{qag}\selectfont \tiny ID}};    \draw[white, fill=white] (-0.0625,0.095) 
    circle [radius=0.007];    \end{tikzpicture}
    \hspace{-2mm}}
\foreach \x in {A, ..., Z}{%
    \expandafter\xdef\csname orcid\x\endcsname{\noexpand\href{https://orcid.org/\csname orcidauthor\x\endcsname}{\noexpand\orcidicon}}
    }

\usepackage[utf8]{inputenc} 
\usepackage[T1]{fontenc}    
\usepackage{hyperref}       
\usepackage{url}            
\usepackage{booktabs}       
\usepackage{amsfonts}       
\usepackage{nicefrac}       
\usepackage{microtype}      
\usepackage{xcolor}         

\usepackage{threeparttable}
\usepackage{multirow}
\usepackage{graphicx}
\usepackage{newtxmath}
\usepackage{rotating}

\usepackage{wrapfig}
\usepackage{ragged2e}
\usepackage{subcaption}

\usepackage{CJKutf8}         

\usepackage{fontawesome5} 
\usepackage{academicons}  
\usepackage{xcolor}       

\usepackage{threeparttable}
\usepackage{multirow}
\usepackage{graphicx}
\usepackage{newtxmath}
\usepackage{rotating}
\usepackage{adjustbox}
\usepackage{wrapfig}
\usepackage{ragged2e}
\usepackage{listings}
\usepackage{ulem} 
\usepackage{arydshln}
\usepackage{makecell}
\usepackage[table]{xcolor}

\usepackage[capitalize,noabbrev]{cleveref}
\usepackage{lipsum}
\usepackage{booktabs}
\usepackage{multirow}
\usepackage{array}
\usepackage{arydshln}
\usepackage{colortbl}
\usepackage{fontawesome5}
\usepackage[utf8]{inputenc}
\usepackage{tabularx}   
\usepackage{amsmath}    
\usepackage{pifont}      
\usepackage{fontawesome5}

\definecolor{tabblue}{HTML}{1f77b4}
\definecolor{taborange}{HTML}{ff7f0e}
\definecolor{tabgreen}{HTML}{2ca02c}
\definecolor{tabred}{HTML}{d62728}
\definecolor{tabpurple}{HTML}{9467bd}
\definecolor{tabpink}{HTML}{ff0080}
\newcolumntype{C}{>{\centering\arraybackslash}X}
\newcommand{\cmark}{\textcolor{green!70!black}{\ding{51}}} 
\newcommand{\xmark}{\textcolor{red!70!black}{\ding{55}}}    
\newcommand{\pub}[1]{$_{\textcolor{gray}{\text{\textit{#1}}}}$}

\hypersetup{
	colorlinks=true,
	linkcolor=red,     
	citecolor=blue,     
	filecolor=blue,     
	urlcolor=blue       
}

\begin{document} 

\title{An Open-Source Benchmark and Baseline for Multi-temporal Referring Segmentation}

\author{
Bingyu~Li*,
Da~Zhang*,
Tao~Huo,
Zhiyuan~Zhao,
Junyu~Gao,~\IEEEmembership{Member,~IEEE,}
~and~
Xuelong~Li,~\IEEEmembership{Fellow,~IEEE}
\thanks{* Equal contribution. Bingyu Li: conceptualization, dataset construction, model framework design, evaluation, and original draft preparation. Da Zhang: conceptualization and manuscript review and editing.}
\thanks{Bingyu Li is with the University of Science and Technology of China, Hefei, China (e-mail: libingyu0205@mail.ustc.edu.cn). Bingyu Li is also with the Institute of Artificial Intelligence (TeleAI), China Telecom, China. This work was done during Bingyu Li's internship at TeleAI.}
\thanks{Da Zhang, Tao Huo, and Junyu Gao are with the School of Artificial Intelligence, Optics and Electronics (iOPEN), Northwestern Polytechnical University, Xi'an 710072, China. Da Zhang, Tao Huo, and Junyu Gao are also with the Institute of Artificial Intelligence (TeleAI), China Telecom, China.}
\thanks{Zhiyuan Zhao and Xuelong Li are with the Institute of Artificial Intelligence (TeleAI), China Telecom, China.}
\thanks{Corresponding authors: Xuelong Li (e-mail: xuelong\_li@ieee.org).}

\href{https://libingyu01.github.io/MTRefSeg/}{\textcolor{black}{\faGithub}  \textbf{\textcolor{tabpink}{Project Page}}}


For any inquiries, please contact \textcolor{tabpink}{\texttt{libingyu0205@mail.ustc.edu.cn}, \texttt{libingyu0205@163.com}.}
}

\maketitle

\begin{abstract}
Large Vision--Language Models (LVLMs) have shown strong visual understanding and language-guided grounding abilities, yet their capacity for multi-temporal visual reasoning remains underexplored. 
To bridge this gap, we introduce \textbf{Multi-temporal Referring Segmentation (MTRS)}, a new task that aims to segment language-described temporal changes from multi-temporal images. 
MTRS extends conventional referring segmentation and change detection by jointly requiring temporal correspondence reasoning, language grounding, and pixel-level mask prediction.
We propose \textbf{CRAFT-Agent}, an automated data construction pipeline with human auditing, and build \textbf{MTRefSeg-21K}, the first MTRS benchmark, containing 21K high-quality multi-temporal image--text--mask triplets across diverse scenes, viewpoints, and domains.
Benchmarking a broad set of VLM- and LVLM-based models reveals that direct inference performs poorly, while task-specific fine-tuning remains limited. 
To address this, we propose \textbf{MTRefSeg-R1}, a change-aware LVLM framework trained with a two-stage strategy.
It first learns general temporal-change perception from 20K vision-only bi-temporal samples, and is then fine-tuned on MTRefSeg-21K for fine-grained language-guided temporal localization.
MTRefSeg-R1 explicitly models cross-temporal visual differences, aligns language instructions with temporal variations, and predicts referred change masks.
Extensive experiments show that MTRefSeg-R1 achieves strong and often superior performance compared with existing LVLM baselines, demonstrating the challenge and potential of MTRS.
\end{abstract}


\input{sec/1_intro}
\input{sec/2_relawork}
\input{sec/3_method}
\input{sec/4_experiment}
\input{sec/5_conclusion}

\section*{Acknowledgements}

The authors would like to thank Chenggang Rong from Northwestern Polytechnical University, 
Du Wu from Northwestern Polytechnical University, 
and Feiyu Wang from Fudan University for their valuable assistance in data cleaning and annotation refinement. 
The authors also thank Haocheng Dong from the University of Science and Technology of China for valuable discussions and support. 
The authors are grateful to Liang Yao from Hohai University for his constructive feedback, which inspired us to reconsider the progression of vision-language segmentation through the lens of multi-temporal visual understanding.

\bibliographystyle{IEEEtran}
\bibliography{reference}

\end{document}

%% file: sec/1_intro.tex
\section{Introduction}
\label{sec:intro}

Large Vision--Language Models (LVLMs) \cite{achiam2023gpt, yang2025qwen3, bai2023qwen} have recently achieved remarkable progress in connecting natural language with visual perception. 
Benefiting from large-scale multimodal pretraining, these models can interpret human instructions, recognize objects, reason about visual scenes, and support language-guided localization or segmentation \cite{wu2025f, zhang2025uwbench, lai2024lisa}. 
Such capabilities are particularly important for fine-grained human--AI interaction in practical applications, including robotics, surveillance, autonomous driving, and remote sensing interpretation \cite{yao2025remotesam, munasinghe2025videoglamm, rasheed2024glamm}.

Recent \textit{AI Flow} studies emphasize interactive and scenario-driven intelligence \cite{an2026ai}, motivating language-guided visual perception to move beyond static understanding toward temporal reasoning in dynamic scenes.

\begin{figure}[t]
\centering
\includegraphics[width=\linewidth]{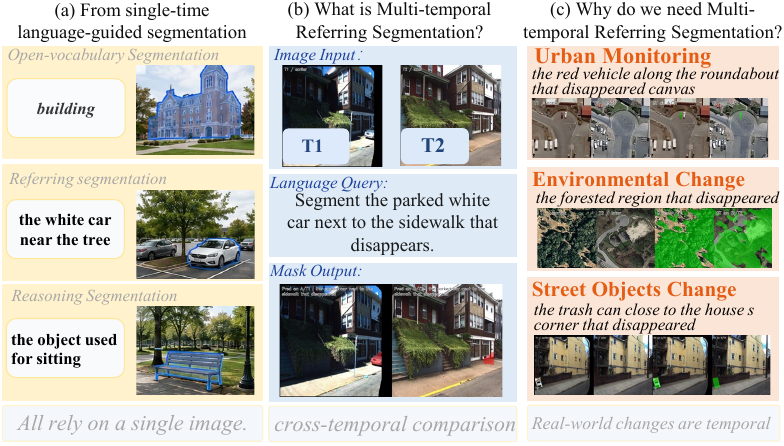}
\caption{
\textbf{Task motivation of Multi-temporal Referring Segmentation (MTRS).}
MTRS addresses this gap by taking temporally related images and a natural-language expression as input, and segmenting the region corresponding to the described temporal change.
}
\label{fig:pic_1_1}
\vspace{-15pt}
\end{figure}

Despite these advances, existing language-guided segmentation tasks mainly focus on single-time visual inputs.
As shown in \cref{fig:pic_1_1}(a), open-vocabulary segmentation typically uses category-level prompts, referring segmentation introduces explicit instance-level descriptions, and reasoning segmentation further requires implicit semantic inference.
However, these tasks still assume that the target can be identified from one static observation alone.
This assumption limits their applicability in real-world scenarios where visual changes unfold over time.
For example, urban monitoring may require locating a vehicle that disappears across two observations, environmental analysis may need to segment a forested region that has vanished, and street-scene understanding may involve identifying a removed object near a specific landmark.
In these cases, the model must not only understand the language query, but also compare temporally related images and determine which region has changed.

To bridge this gap, we introduce \textbf{Multi-temporal Referring Segmentation (MTRS)}, a new language-guided segmentation task that advances segmentation from single-time perception to bi-temporal and multi-temporal visual reasoning.
Given a set of temporally related images and a natural language expression, MTRS requires the model to identify and segment the region corresponding to the described temporal variation.
In this paper, we mainly instantiate the task in the bi-temporal setting, which is the most common formulation in change detection, while keeping the task definition compatible with more general multi-temporal inputs.
Compared with conventional referring image segmentation, MTRS introduces an additional temporal reasoning requirement: the target cannot be reliably determined from a single image, but must be inferred through cross-temporal comparison.
Compared with traditional change detection, MTRS is more flexible and interactive, since the target region is specified by natural language rather than by a fixed category label or a binary change map.

\begin{figure*}[t]
\centering
\includegraphics[width=\linewidth]{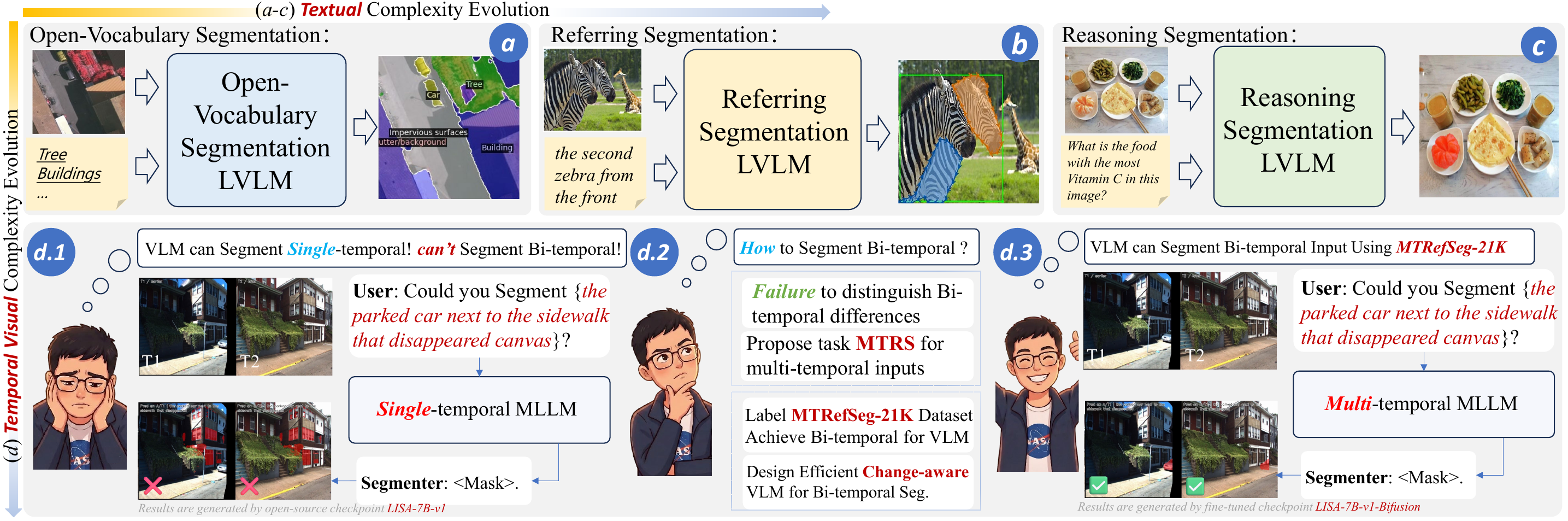}
\caption{
\textbf{From single-time language-guided segmentation to multi-temporal referring segmentation.}
Complementary to the task motivation in \cref{fig:pic_1_1}, this figure summarizes the evolution of segmentation tasks from both textual and visual complexity.
From the textual perspective, language-guided segmentation evolves from category-level open-vocabulary segmentation, to instance-level referring segmentation, and further to reasoning segmentation with implicit semantic understanding.
From the visual perspective, MTRS introduces an additional temporal dimension, requiring models to compare temporally related images and segment the language-specified change rather than a static object.
This motivates the construction of \textbf{MTRefSeg-21K} and the design of a change-aware LVLM framework for language-grounded temporal change segmentation.
}
\label{fig:pic_1}
\vspace{-10pt}
\end{figure*}

Figure~\ref{fig:pic_1_1} provides an intuitive motivation for MTRS by contrasting conventional single-time language-guided segmentation with temporal-change-oriented referring segmentation.
It shows that the key difficulty of MTRS does not merely lie in recognizing objects or following textual descriptions, but in identifying which region is temporally changed and simultaneously relevant to the given language query.
Building upon this task-level motivation, \cref{fig:pic_1} further presents the broader evolution of language-guided segmentation and the systematic formulation of our benchmark and framework.
Specifically, while \cref{fig:pic_1_1} answers \textit{what MTRS is} and \textit{why it is needed}, \cref{fig:pic_1} explains \textit{how MTRS extends existing segmentation paradigms} and \textit{how our work supports this new task}.

This new task poses several unique challenges.
First, models need to establish reliable temporal correspondence between images that may contain viewpoint changes, illumination variation, seasonal differences, or object displacement.
Second, they must distinguish language-relevant temporal changes from irrelevant visual differences.
For instance, a query such as \textit{``the newly placed vehicle near the building''} requires the model to focus on a specific changed object rather than all changed regions.
Third, the model must generate accurate pixel-level masks for the referred temporal region, requiring both high-level semantic reasoning and low-level spatial precision.

To facilitate systematic study of MTRS, we construct \textbf{MTRefSeg-21K}, a large-scale benchmark for multi-temporal referring segmentation.
As illustrated in \cref{fig:craft_agent}, we design a Cross-temporal Referring Automated Formation and Refinement Agent, termed \textbf{CRAFT-Agent}, to automatically generate high-quality multi-temporal bi-image--text--mask annotations with human auditing.
The pipeline first identifies candidate temporal changes from image pairs, then generates spatially grounded referring expressions, refines segmentation masks, and finally rewrites mechanical descriptions into natural language expressions.
Using this pipeline, MTRefSeg-21K provides 21K high-quality bi-image--text--mask triplets across diverse viewpoints and domains, including aerial-view scenes, remote sensing imagery, and normal-view environments.
Each sample contains temporally related images, a natural language expression describing a target temporal change, and the corresponding segmentation mask.

Beyond dataset construction, MTRefSeg-21K supports comprehensive evaluation under the overall, NS-domain, and RS-domain settings, enabling systematic analysis of both general-scene and remote-sensing multi-temporal referring segmentation.
Our benchmark reveals that existing LVLMs still struggle with multi-temporal referring segmentation.
Direct inference often fails to accurately distinguish temporal differences or localize the language-specified region.
Although task-specific fine-tuning improves temporal reasoning and mask prediction, the performance remains far from satisfactory.
These observations suggest that single-temporal visual-language pretraining is insufficient for MTRS, since the model must first acquire general temporal-change perception before learning fine-grained language-guided localization.

To address these limitations, we propose \textbf{MTRefSeg-R1}, a change-aware LVLM framework with a two-stage training strategy for multi-temporal referring segmentation.
Rather than directly performing end-to-end multimodal pretraining with large-scale image--text annotations, MTRefSeg-R1 first learns general temporal-change perception from approximately 20K vision-only bi-temporal change samples.
This Stage-1 visual pretraining focuses on cross-temporal comparison and change localization, allowing the model to acquire change-sensitive visual representations before being exposed to language instructions.
In the second stage, the model is fine-tuned on MTRefSeg-21K with fine-grained referring expressions and pixel-level masks, thereby strengthening its ability to align natural-language instructions with specific temporal-change regions.

Architecturally, MTRefSeg-R1 explicitly models temporal discrimination by comparing temporally correlated images, aligning language instructions with temporal variations, and predicting the referred change masks.
Instead of treating multi-temporal images as independent inputs, the framework emphasizes change-aware temporal reasoning and language-conditioned segmentation.
By integrating vision-only temporal-change pretraining, referring segmentation fine-tuning, and change-aware model design, MTRefSeg-R1 achieves stronger performance than existing LVLM baselines across multiple evaluation settings.

In summary, our contributions are four-fold:

\begin{itemize}
    \item We introduce \textbf{Multi-temporal Referring Segmentation (MTRS)}, a new task that unifies temporal change understanding, natural language grounding, and pixel-level segmentation.
    \item We propose \textbf{CRAFT-Agent}, an automated data construction pipeline with human auditing, and build \textbf{MTRefSeg-21K}, the first large-scale benchmark for MTRS, containing 21K high-quality bi-image--text--mask triplets across diverse scenes, viewpoints, and domains.
    \item We design a two-stage training strategy for MTRS, where the model first learns change-sensitive visual representations from approximately 20K vision-only bi-temporal change samples and is then fine-tuned with language-guided mask supervision on MTRefSeg-21K.
    \item We provide comprehensive benchmarking of existing LVLMs and propose \textbf{MTRefSeg-R1}, a change-aware LVLM framework that substantially improves multi-temporal referring segmentation performance.
\end{itemize}

%% file: sec/2_relawork.tex
\section{Related Work}

\subsection{Large Vision--Language Models (LVLMs)}

LVLMs have achieved remarkable progress in recent years \cite{bai2023qwen, yang2025qwen3, achiam2023gpt, xu2025stare, shen2025fine, li2026toward, shen2026egoforge, sarkar2025reasoning, palikhe2025towards, yu2025yielding, yu2025cotextor, yu2025forgetme, meng2026tri, chen2026generative}. 
Recent studies have demonstrated that LVLMs can effectively bridge high-level semantic reasoning with low-level visual grounding \cite{rasheed2024glamm, wu2025f, zhang2025uwbench, danish2025geobench, sarkar2025reasoning, meng2026tri}. 
This capability makes them particularly suitable for tasks that require both comprehensive scene understanding and precise spatial localization. 
Beyond general multimodal reasoning, recent works have further explored the application of LVLMs to perception-oriented tasks such as object detection, language-guided localization, and segmentation \cite{ma2025geomag, shabbir2025geopixel, munasinghe2025videoglamm, yu2025yielding, yu2026dinov3}. 
By integrating LVLMs with foundation models such as the Segment Anything Model (SAM) and grounding-based detectors \cite{yao2025remotesam, yu2026dinov3}, these approaches enable open-vocabulary and interactive visual understanding. 
In such frameworks, LVLMs typically serve as the semantic reasoning module that interprets user instructions, while specialized perception models generate accurate spatial predictions. 
This design allows the system to combine the flexibility of natural language understanding with the precision of dedicated visual backbones \cite{yu2025cotextor, palikhe2025towards}. 

However, most existing LVLM-based perception systems are designed for \textit{single-time} images, which limits their ability to capture temporal correspondence across multi-temporal visual inputs. 
Such a limitation becomes particularly significant in scenarios involving temporal dynamics, cross-temporal comparison, semantic change understanding, or long-term scene monitoring \cite{yu2025visualizing, yu2026spatiotemporal}. 
In this work, we explore whether LVLMs can effectively reason over multi-temporal visual inputs and segment the regions corresponding to language-described temporal variations \cite{sarkar2025reasoning, yu2025visualizing}. 
This setting requires the model to jointly understand temporal changes, natural language expressions, and pixel-level spatial grounding.

\begin{figure*}
\centering
\includegraphics[width=\linewidth]{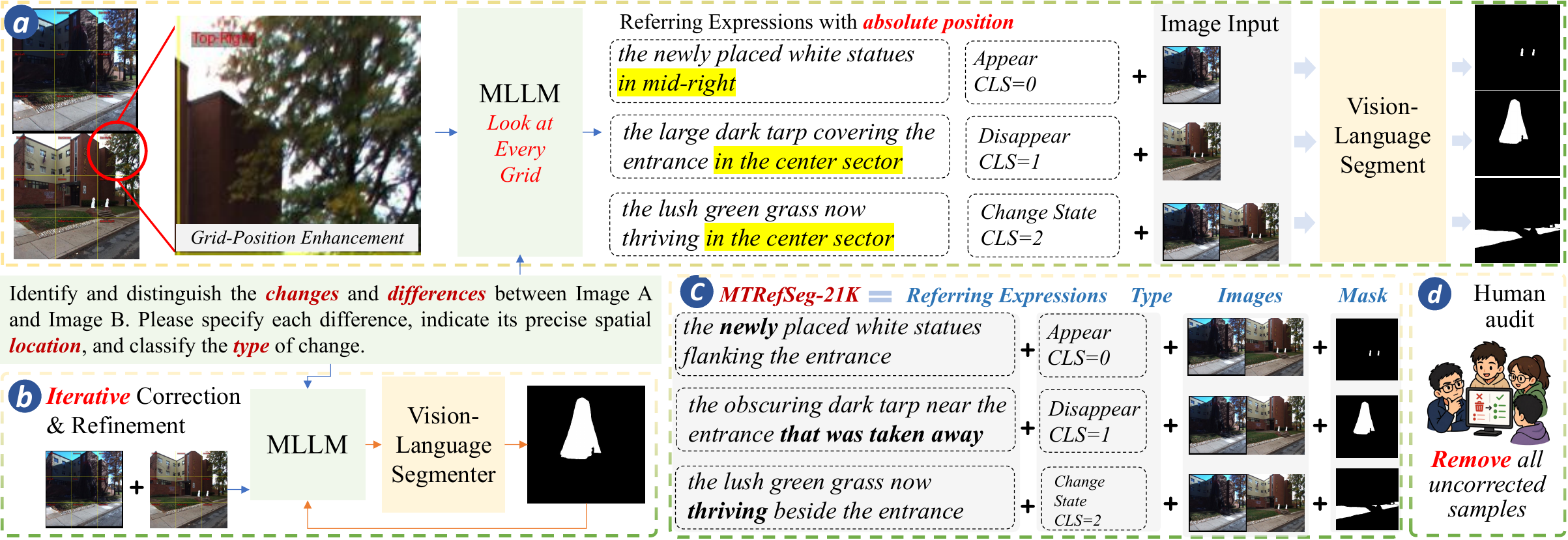}
\caption{
\textbf{Overview of CRAFT-Agent.}
CRAFT-Agent constructs bi-image--text--mask triplets for MTRS by integrating grid-aware change perception, expression generation, mask refinement, expression beautification, and human auditing.
}
\vspace{-5pt}
\label{fig:craft_agent}
\end{figure*}

\subsection{Language-guided Localization and Segmentation}

As summarized in \cref{fig:pic_1}(a--c), existing language-guided localization and segmentation tasks can be broadly categorized into three paradigms: \textit{Open-Vocabulary}, \textit{Referring}, and \textit{Reasoning}. 
These paradigms mainly differ in the complexity and expressiveness of textual inputs, progressing from category-level prompts, to explicit instance descriptions, and further to implicit reasoning queries \cite{sarkar2025reasoning, palikhe2025towards, meng2026tri}.

\textbf{Open-Vocabulary}-based tasks aim to segment or detect objects whose categories are defined by a predefined or open vocabulary. 
Representative approaches are typically built upon DETR-style architectures \cite{li2025maris, yu2023convolutions}, diffusion-based methods \cite{karazija2024diffusion}, cost aggregation frameworks \cite{cho2024cat, li2025exploring_uw,li2025exploring}, or segmentation pipelines leveraging foundation models such as SAM \cite{zhang2023simple, ding2022open}. 
These methods mainly focus on aligning visual representations with category-level language representations, thereby enabling recognition beyond a closed-set label space.

\textbf{Referring}-based tasks focus on segmenting or localizing a specific instance described by a more detailed natural language expression \cite{liu2023polyformer, wu2024toward, liu2024rotated, shah2024lqmformer, shen2025survey}. 
Compared with open-vocabulary settings, referring tasks enable more precise human--machine interaction by grounding fine-grained textual descriptions in visual scenes \cite{yue2024adaptive, chen2025rsrefseg1, yang2025referring, dai2025deris, yang2022lavt, liu2023gres, chng2024mask}. 
For example, the model may be asked to segment \textit{``the small dog on the left''} rather than simply identifying the category \textit{dog}. 
This setting requires not only category recognition but also instance-level discrimination guided by language.

\textbf{Reasoning}-based tasks further extend this paradigm by requiring models to interpret implicit textual descriptions and localize the target object accordingly \cite{lai2024lisa, wang2024segllm, shen2025reasoning, sarkar2025reasoning}. 
Such tasks typically rely on LVLMs and emphasize the model's capability for \textit{implicit reasoning} \cite{liu2025seg, yang2023lisa++, palikhe2025towards}. 
Instead of relying solely on explicit object attributes, the model must infer the intended target from more abstract or relational descriptions, making the task substantially more challenging. 

Different from the above settings, this paper studies \textbf{Multi-temporal Referring Segmentation (MTRS)}, where the model receives multi-temporal visual inputs and a natural-language expression, and is required to segment the region corresponding to the described temporal variation. 
As shown in \cref{fig:pic_1}(d.1--d.3), MTRS introduces a new dimension of visual complexity by moving from single-time segmentation to bi-temporal change-aware segmentation. 
Therefore, the referred target is defined not only by its category, attributes, or spatial relations, but also by its temporal state and cross-time variation \cite{yu2025visualizing, yu2026spatiotemporal}.

\subsection{Multi-temporal Change Detection and Segmentation}

Multi-temporal image models are widely studied in \textit{change detection} tasks \cite{radolko2016dataset, wang2025mds, yu2025qrs}, where the most common formulation is bi-temporal change detection. 
These methods typically take two images captured at different times as input and predict whether each pixel has changed. 
Although some approaches extend this formulation to multi-class change detection or semantic change segmentation, they generally rely on fixed and predefined label sets, which limits their flexibility and generalization ability in open-world scenarios. 
Moreover, most traditional change detection methods do not support natural language input, making human--machine interaction less intuitive and less flexible \cite{yu2025cotextor}. 

Recent efforts have explored \textit{open-vocabulary} or language-aware change detection using vision foundation models \cite{li2025dynamicearth, zhu2025semantic, zhang2025unichange, yu2026dinov3}. 
These methods improve the flexibility of change analysis by introducing category-level textual prompts or open-vocabulary recognition. 
However, many of them are still based on pipelines that combine relatively small perception models, such as SAM and Grounding-DINO, rather than fully leveraging the unified perception and reasoning capabilities of LVLMs. 
As a result, their ability to support interactive, description-driven, and language-guided temporal segmentation remains limited \cite{palikhe2025towards, yu2025yielding}. 

In particular, although recent work has also introduced the term \textit{Referring Change Detection} \cite{korkmaz2026referring}, its formulation remains closer to category-conditioned change detection than to genuine description-driven multi-temporal referring segmentation. 
Specifically, the text input is mainly used to specify a target change category, and the model predicts the binary change region associated with that category. 
Such a formulation remains closely aligned with conventional semantic or binary change detection, where language primarily functions as a class selector rather than a rich semantic description of the temporal event itself. 
Consequently, it provides limited support for more expressive linguistic phenomena, such as object-level attributes, spatial relations, temporal states, and fine-grained semantic modification. 

By contrast, our MTRS setting emphasizes language-guided segmentation over multi-temporal visual inputs.
Instead of merely selecting a predefined change category, the model is expected to interpret richer descriptions of temporal variation and identify the corresponding regions under flexible natural language instructions. 
Furthermore, our \textbf{MTRefSeg-21K} benchmark covers a broader range of scenes and supports more diverse and semantically expressive annotations, making the task more challenging and better aligned with realistic human--machine interaction. 
In contrast, this work focuses on a setting that integrates multi-temporal visual perception with language-guided reasoning.
Our goal is to examine whether LVLMs can effectively perceive and segment language-described temporal changes, thereby enabling a more flexible, expressive, and interactive form of multi-temporal visual understanding \cite{deng2025changechat, zhou2025segchange, lu2023viewpoint, korkmaz2026referring, yu2025visualizing, yu2026spatiotemporal}.

%% file: sec/3_method.tex
\section{MTRefSeg-21K Dataset Construction}
\label{mtrefseg}

\begin{figure*}[t]
\centering
\includegraphics[width=0.95\linewidth]{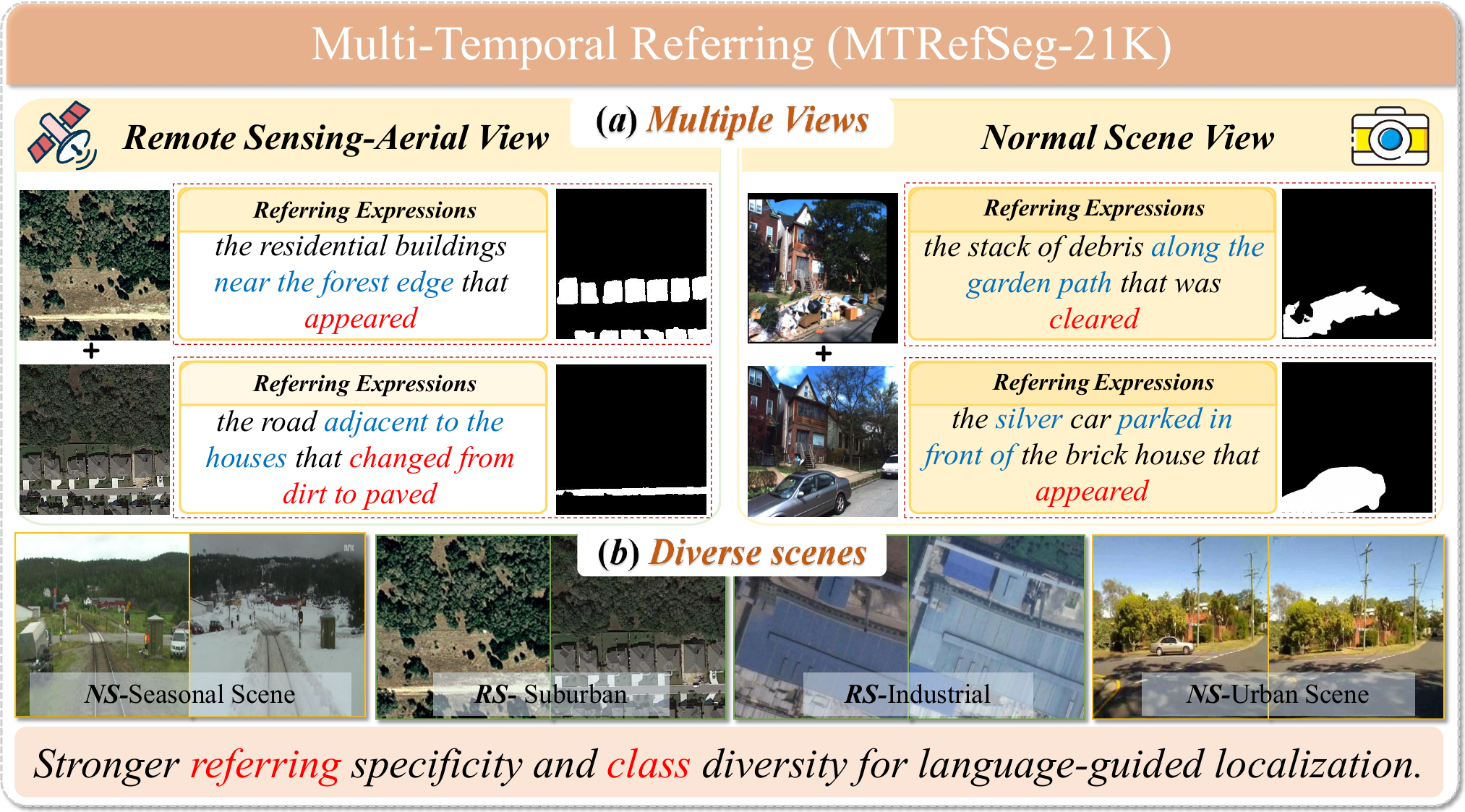}
\caption{
\textbf{Data design for MTRS.}
\textbf{MTRefSeg-21K} provides fine-grained bi-image--text--mask triplets for language-guided multi-temporal referring fine-tuning across RS and NS domains.
}
\label{fig:pic_dataset}
\vspace{-1em}
\end{figure*}

\subsection{CRAFT-Agent}

To enable scalable construction of multi-temporal referring segmentation annotations, we design a Cross-temporal Referring Automated Formation and Refinement Agent (CRAFT-Agent), illustrated in \cref{fig:craft_agent}. 
The agent automatically generates high-quality bi-image--text--mask triplets for \textbf{Multi-temporal Referring Segmentation (MTRS)} through a three-stage pipeline. 
First, a grid-aware multimodal large language model examines bi-temporal image pairs and identifies differences across spatial regions, producing referring expressions with explicit spatial cues and temporal variation categories (e.g., appear, disappear, or state change) together with initial segmentation masks. 
Second, an \underline{iterative refinement} process progressively improves mask boundary accuracy through a vision-language segmentation model. 
Finally, an expression beautification module rewrites the mechanically generated spatial descriptions into natural and contextually coherent referring expressions. 
This automated pipeline produces high-quality bi-image--text--mask annotations that serve as the foundation for constructing \textbf{MTRefSeg-21K}.

\subsection{MTRefSeg-21K}

Based on the proposed CRAFT-Agent, we construct \textbf{MTRefSeg-21K}, a benchmark for \textbf{Multi-temporal Referring Segmentation (MTRS)} containing 9,521 bi-temporal image pairs and 20,924 referring expressions across diverse viewpoints, including aerial-view scenes, remote sensing imagery, and normal-view environments, as shown in Fig. \ref{fig:pic_dataset}. 
Each sample consists of a pair of temporally related images, one or more referring expressions describing specific object-level temporal changes, and the corresponding segmentation masks. 
The dataset follows a multi-view design and covers diverse scenes such as industrial areas, suburban regions, seasonal environments, urban scenes, and traffic scenarios. 
In addition to the standard supervised setting, MTRefSeg-21K also supports cross-domain evaluation between Remote Sensing (RS) and Normal Scene (NS) domains, providing a comprehensive benchmark for evaluating language-guided temporal change understanding and cross-domain generalization.

\subsubsection{Dataset Split Analysis}

Fig.~\ref{fig:three_images} and Table~\ref{tab:combined_dataset_tables} summarize the key properties of \textit{MTRefSeg-21K}. 
Our dataset contains 9,521 bi-temporal image pairs and 20,924 referring expressions, with 5,646/3,875 image pairs and 12,438/8,486 expressions in the train/val split. 
It covers both Normal Scene (NS) and Remote Sensing (RS) domains, including 1,396 NS image pairs with 4,654 expressions and 8,125 RS image pairs with 16,270 expressions, enabling multi-domain evaluation within a unified benchmark.

\subsubsection{Dataset Distributions Analysis}

As shown in Fig.~\ref{fig:three_images}, the mask-area distributions across different splits are highly consistent and exhibit a clear long-tailed pattern, indicating that \textit{MTRefSeg-21K} covers temporal changes of diverse scales, from small local edits to large structural variations. 
The expression-length distributions are also stable across splits, with most descriptions containing 8--12 words, suggesting concise yet informative language annotations. 
The word cloud further shows that the dataset contains rich object-level semantics and spatial relation cues, such as \textit{building}, \textit{road}, \textit{adjacent}, and \textit{beside}, which are critical for multi-temporal referring segmentation.

\subsubsection{Comparison with Other Datasets}

Table~\ref{tab:combined_dataset_tables} compares \textit{MTRefSeg-21K} with existing general-domain and remote-sensing referring segmentation benchmarks. 
Unlike conventional referring segmentation datasets, \textit{MTRefSeg-21K} explicitly models bi-temporal changes. 
Compared with existing remote-sensing datasets, it provides richer language annotations, covers both Normal Scene and Remote Sensing domains, and supports a wider resolution range from $224^2$ to $2048^2$. 
More importantly, it unifies general-domain and remote-sensing data in a single benchmark. 
These properties make \textit{MTRefSeg-21K} a diverse and uniquely suitable benchmark for \textbf{Multi-temporal Referring Segmentation (MTRS)}.

\input{table/1_datasetAnalysis_0}

\input{table/1_datasetAnalysis_1}

\begin{table}[t]
\centering
\caption{Category diversity of MTRefSeg-21K across domains.}
\label{tab:domain_category_statistics}
\setlength{\tabcolsep}{5pt}
\renewcommand{\arraystretch}{1.08}
\begin{tabular}{lccc}
\toprule
\textbf{Domain} 
& \textbf{Expr.} 
& \textbf{Fine-grained} 
& \textbf{Coarse} \\
\midrule
Normal Scene (NS)   & 4,654  & 296 & 9 \\
Remote Sensing (RS) & 16,270 & 254 & 7 \\
\midrule
Union               & 20,924 & 442 & 9 \\
Intersection        & --     & 108 & 7 \\
\bottomrule
\end{tabular}
\vspace{-1em}
\end{table}

\begin{table}[t]
\centering
\caption{Coarse-category distribution of MTRefSeg-21K across NS and RS domains. Each cell reports count / ratio.}
\label{tab:coarse_category_distribution}
\setlength{\tabcolsep}{4.5pt}
\renewcommand{\arraystretch}{1.08}
\resizebox{\linewidth}{!}{
\begin{tabular}{llll}
\toprule
\textbf{Coarse Category} 
& \textbf{NS} 
& \textbf{RS} 
& \textbf{Overall} \\
\midrule
Building-related      & 135 / 2.90\%    & 10,557 / 64.89\% & 10,692 / 51.10\% \\
Other                 & 2,495 / 53.61\% & 2,232 / 13.72\%  & 4,727 / 22.59\%  \\
Vehicle               & 815 / 17.51\%   & 872 / 5.36\%     & 1,687 / 8.06\%   \\
Road surface          & 138 / 2.97\%    & 1,526 / 9.38\%   & 1,664 / 7.95\%   \\
Vegetation/open space & 173 / 3.72\%    & 985 / 6.05\%     & 1,158 / 5.53\%   \\
Waste cleanup         & 412 / 8.85\%    & 14 / 0.09\%      & 426 / 2.04\%     \\
Construction/infra.   & 262 / 5.63\%    & 84 / 0.52\%      & 346 / 1.65\%     \\
Sign/marking          & 151 / 3.24\%    & 0 / 0.00\%       & 151 / 0.72\%     \\
Other scene object    & 73 / 1.57\%     & 0 / 0.00\%       & 73 / 0.35\%      \\
\bottomrule
\end{tabular}
}
\vspace{-2em}
\end{table}

\subsubsection{Category Distribution Analysis}

Tables~\ref{tab:domain_category_statistics} and~\ref{tab:coarse_category_distribution} summarize the category composition of MTRefSeg-21K across the NS and RS domains.
Although RS contains more expression instances, NS covers more fine-grained categories, indicating richer object-level diversity in normal scenes.
The union and intersection statistics show that the dataset includes both shared cross-domain semantics and domain-specific categories, supporting evaluation under category and domain shifts.
At the coarse level, RS is dominated by building-related changes, whereas NS shows a more diverse distribution involving vehicles, waste cleanup, and construction/infrastructure.
These statistics demonstrate that MTRefSeg-21K provides a diverse and challenging benchmark for language-guided temporal change segmentation across heterogeneous visual domains.

\section{Method}
\label{sec:method}

\subsection{Problem Definition}

We define \textbf{Multi-temporal Referring Segmentation (MTRS)} as a language-guided bi-temporal segmentation task. Although MTRS can be naturally extended to more than two timestamps, this paper focuses on the bi-temporal case as the fundamental and most widely used setting.
Given an earlier image $\mathbf{I}^{t_1}$, a later image $\mathbf{I}^{t_2}$, and a natural-language expression $\mathbf{x}$ describing a target temporal change, the goal is to predict a binary mask $\mathbf{M}$ corresponding to the referred changed region.

Formally, the input is defined as:
\begin{equation}
\label{eq:mtrs_input}
(\mathbf{I}^{t_1}, \mathbf{I}^{t_2}, \mathbf{x}),
\end{equation}
where $\mathbf{I}^{t_1}$ and $\mathbf{I}^{t_2}$ denote two temporally separated observations of the same scene, and $\mathbf{x}$ specifies the target change through expressions such as ``appeared'', ``disappeared'', ``newly placed'', or ``removed''. 
The objective is to learn a mapping:
\begin{equation}
\label{eq:mtrs_mapping}
f_{\theta}(\mathbf{I}^{t_1}, \mathbf{I}^{t_2}, \mathbf{x}) = \mathbf{M},
\end{equation}
where $\theta$ denotes the model parameters and $\mathbf{M} \in \{0,1\}^{H \times W}$ is the segmentation mask of the referred temporal-change region.

Compared with conventional referring image segmentation, MTRS requires cross-temporal comparison rather than single-image grounding. 
Compared with traditional change detection, MTRS does not aim to detect all changed regions, but instead localizes only the region specified by the language instruction. 
Therefore, MTRS jointly requires temporal correspondence reasoning, language-conditioned change understanding, and pixel-level mask prediction.

Given a training set:
\begin{equation}
\label{eq:mtrs_dataset}
\mathcal{D} = \left\{(\mathbf{I}^{t_1}_i, \mathbf{I}^{t_2}_i, \mathbf{x}_i, \mathbf{M}_i)\right\}_{i=1}^{N},
\end{equation}
the model is optimized to generate masks that are consistent with both the bi-temporal visual evidence and the referring expression.

\subsection{Generalizing Existing Models to MTRS}
\label{sec:adapt_mtrs}

Existing vision-language segmentation models are mainly designed for single-time images and cannot directly handle language-described temporal changes. 
To provide strong baselines for MTRefSeg-21K, we extend representative VLMs and LVLMs to the MTRS setting by introducing paired image inputs, temporal feature interaction, and temporally grounded instructions.

\subsubsection{Adapting VLM-based Segmentation Models}

For conventional VLM-based referring segmentation models, we replace the original single-image input with a pair of temporally ordered images. 
Given $\mathbf{I}^{t_1}$ and $\mathbf{I}^{t_2}$, the two images are encoded by a shared visual encoder:
\begin{equation}
\label{eq:vlm_visual_encode}
\mathbf{F}^{t_1} = E_v(\mathbf{I}^{t_1}), \qquad
\mathbf{F}^{t_2} = E_v(\mathbf{I}^{t_2}),
\end{equation}
where $E_v(\cdot)$ denotes the visual backbone. 
The shared encoder preserves semantic consistency between the two temporal observations while avoiding additional model complexity.

The extracted features are then fused into a change-aware representation:
\begin{equation}
\label{eq:vlm_fusion}
\mathbf{F}^{\Delta} = \Phi(\mathbf{F}^{t_1}, \mathbf{F}^{t_2}),
\end{equation}
where $\Phi(\cdot)$ denotes a temporal fusion operator, such as feature concatenation, differencing, additive fusion, or a lightweight learnable fusion block. 
The fused representation $\mathbf{F}^{\Delta}$ is fed into the segmentation decoder together with the text representation to predict the referred change mask. 
The language branch remains unchanged, where the referring expression is encoded by the original text encoder, such as BERT \cite{devlin2019bert} or the CLIP text encoder \cite{radford2021clip}. 
This strategy enables existing VLM-based models to support MTRS with minimal architectural modification.

\begin{figure}
\centering
\includegraphics[width=\linewidth]{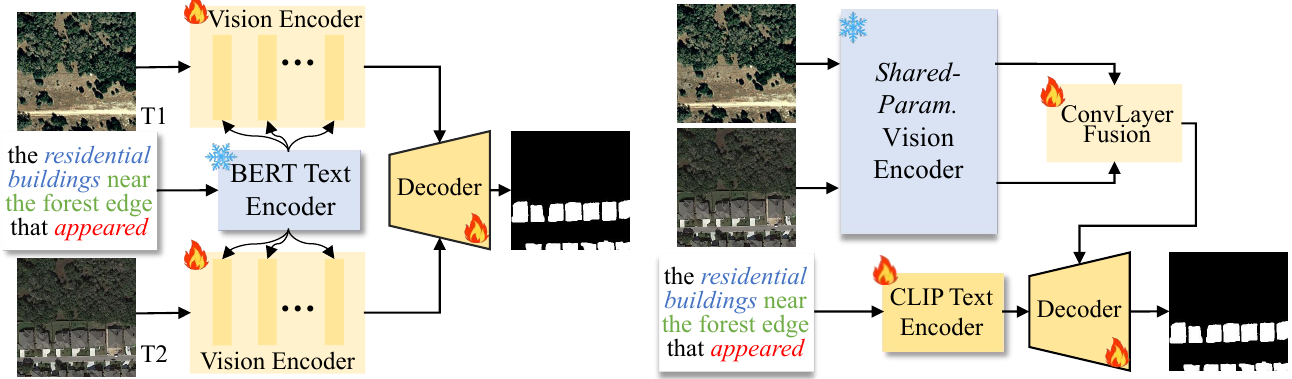}
\caption{
\textbf{Adapting VLM-based segmentation models to MTRS.}
Single-time VLM frameworks are extended with paired image inputs and temporal feature interaction to localize language-described changes.
}
\label{fig:pic_model_modify_vlm}
\vspace{-1em}
\end{figure}

\subsubsection{Adapting LVLM-based Segmentation Models}

For segmentation-oriented LVLMs, supporting MTRS requires both bi-temporal visual modeling and temporally grounded instruction following. 
We first encode the two temporal images using a shared vision encoder:
\begin{equation}
\label{eq:lvlm_shared_encoder}
\mathbf{Z}^{t_1} = E_v(\mathbf{I}^{t_1}), \qquad
\mathbf{Z}^{t_2} = E_v(\mathbf{I}^{t_2}),
\end{equation}
and fuse them into a unified temporal representation:
\begin{equation}
\label{eq:lvlm_fusion}
\mathbf{Z}^{\Delta} = \Psi(\mathbf{Z}^{t_1}, \mathbf{Z}^{t_2}),
\end{equation}
where $\Psi(\cdot)$ denotes the temporal fusion function.

To explicitly encode temporal order, we reformulate the instruction as:
\begin{quote}
\small
\noindent\textbf{Prompt:}
\texttt{``<image\_t1> is the earlier image, and <image\_t2> is the later image. Compare the two images and segment the <Referring Instruction>.''}
\end{quote}
This prompt informs the LVLM that the two images correspond to the same scene observed at different times, and that the output should be conditioned on temporal changes rather than static appearance.

For segmentation-oriented LVLMs, the fused visual representation and the reformulated prompt are jointly processed by the multimodal LLM. 
The hidden state of the segmentation token $[\mathrm{SEG}]$ is then used to guide mask prediction:
\begin{equation}
\label{eq:lvlm_mask}
\mathbf{M} = D_m(\mathbf{Z}^{\Delta}, \mathbf{h}_{[\mathrm{SEG}]}),
\end{equation}
where $D_m(\cdot)$ denotes the mask decoder and $\mathbf{h}_{[\mathrm{SEG}]}$ is the hidden embedding of the segmentation token. 
This adaptation preserves the language reasoning ability of the original LVLM while equipping it with change-aware dense prediction capability.

\begin{figure}
\centering
\includegraphics[width=\linewidth]{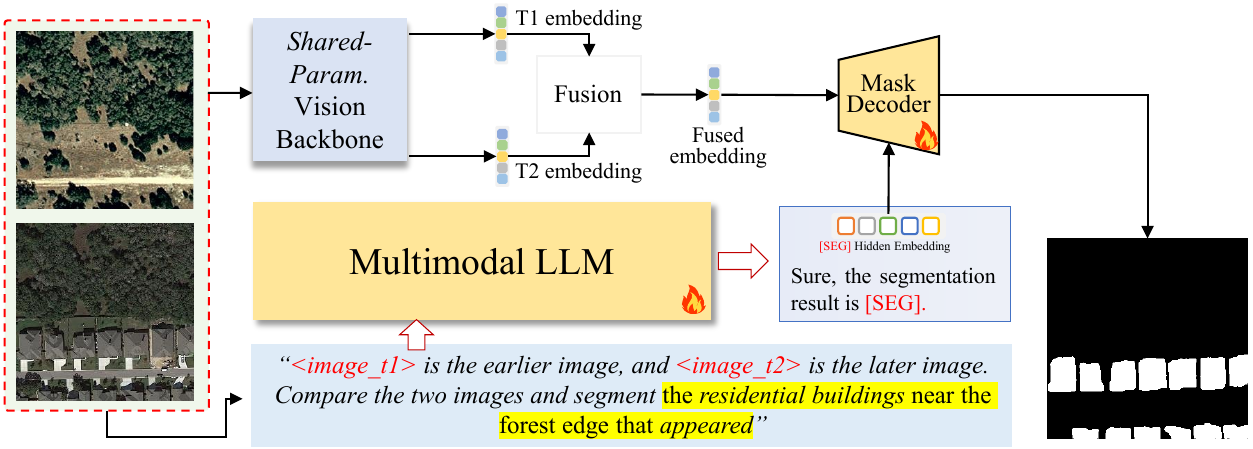}
\caption{
\textbf{Adapting LVLM-based segmentation models to MTRS.}
A segmentation-oriented LVLM is modified to jointly process multi-temporal images and generate masks conditioned on temporal-change descriptions.
}
\label{fig:pic_model_modify_lvlm}
\vspace{-1em}
\end{figure}

\begin{figure*}[t]
\centering
\includegraphics[width=\linewidth]{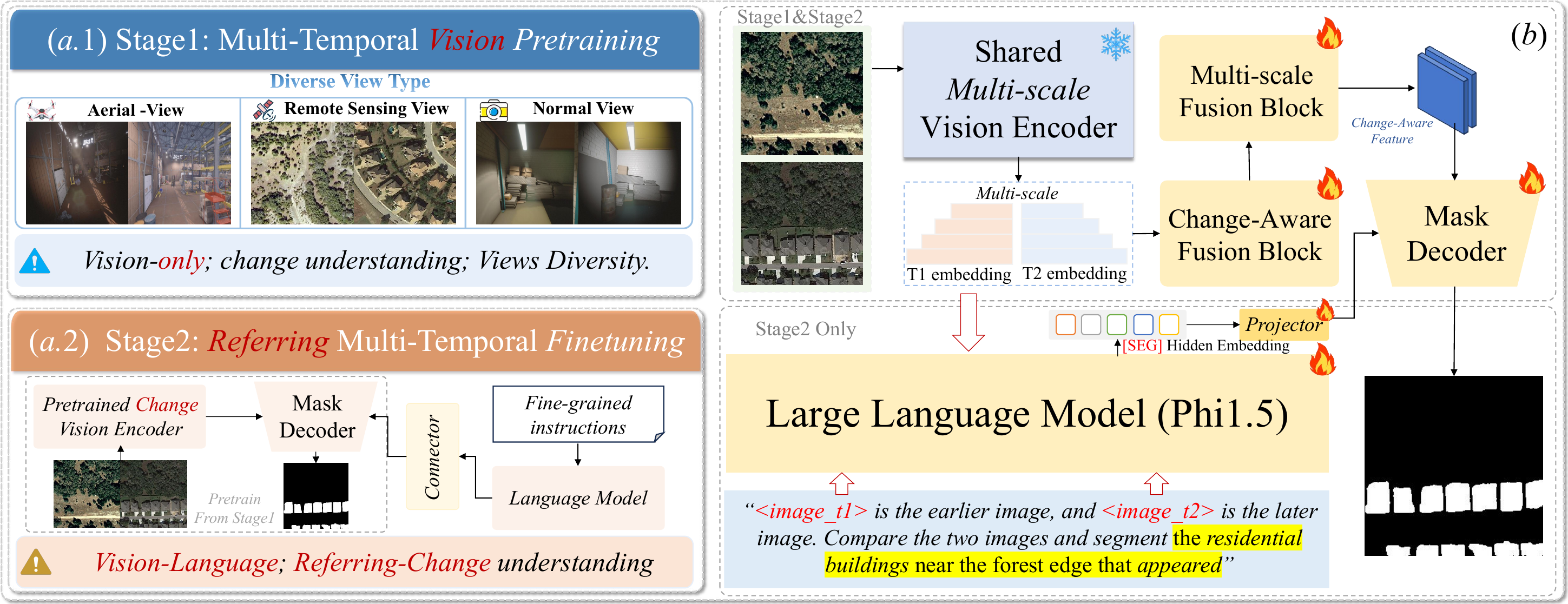}
\caption{
\textbf{Overview of the proposed MTRefSeg-R1 framework.}
MTRefSeg-R1 adopts a two-stage training strategy.
Stage 1 performs multi-temporal vision pretraining on diverse view types to learn generic change-aware visual representations.
Stage 2 conducts referring multi-temporal fine-tuning, where the LVLM understands temporally ordered image pairs and language instructions, and predicts the mask of the referred changed region through the $[\mathrm{SEG}]$ token and the change-aware mask decoder.
}
\label{fig:method}
\vspace{-1em}
\end{figure*}

\subsection{MTRefSeg-R1: Change-Aware LVLM for MTRS}
\label{sec:mtrefseg_r1}

To address Multi-Temporal Referring Segmentation (MTRS), we propose \textbf{MTRefSeg-R1}, a change-aware large vision-language model that jointly performs temporal reasoning, referring expression understanding, and pixel-level changed-region segmentation.
Given an earlier image $\mathbf{I}^{t_1}$, a later image $\mathbf{I}^{t_2}$, and a referring expression $\mathbf{x}$, the model predicts a binary mask $\mathbf{M}$ for the language-specified changed region.
As shown in Fig.~\ref{fig:method}, MTRefSeg-R1 is built upon a two-stage training framework.
The first stage learns generic multi-temporal change perception from vision-only supervision, while the second stage injects language instructions and optimizes the model for referring change segmentation.

\subsubsection{Two-Stage Training Strategy}

MTRefSeg-R1 follows a progressive training strategy to bridge generic temporal-change understanding and fine-grained language-guided mask prediction.

\textbf{Stage 1: Multi-temporal vision pretraining.}
In the first stage, we train the visual change perception branch using vision-only bi-temporal change data.
Specifically, we collect approximately 20K bi-temporal image pairs with binary change masks from existing change detection datasets.
These samples do not contain referring expressions and are used only to supervise generic changed-region localization.
The collected data cover diverse view types, including aerial-view scenes, remote-sensing imagery, and normal-view environments.
Given a bi-temporal image pair $(\mathbf{I}^{t_1}, \mathbf{I}^{t_2})$, the shared multi-scale vision encoder extracts temporal features from the two images, and the change-aware fusion module aggregates them into multi-scale change representations.
The mask decoder is then optimized to predict the binary changed regions from the fused visual features.
Through this vision-only pretraining stage, the model learns basic temporal correspondence, change localization, and view-robust visual representations before being further adapted to fine-grained language-guided temporal grounding.

\textbf{Stage 2: Referring multi-temporal fine-tuning.}
In the second stage, we fine-tune the model using bi-temporal image--text--mask triplets.
Each sample contains an earlier image, a later image, a referring expression describing a specific temporal change, and the corresponding binary mask.
Different from Stage 1, this stage requires the model to identify not all changed regions, but only the changed object or area specified by the language instruction.
Therefore, we introduce an LVLM reasoning branch to process temporally ordered image tokens and textual instructions.
The hidden state of the generated $[\mathrm{SEG}]$ token is projected as a language-conditioned segmentation query, which guides the mask decoder to output the referred changed region.
This two-stage design allows MTRefSeg-R1 to first acquire general change perception and then learn fine-grained language-conditioned temporal grounding.

\subsubsection{Overall Framework}

MTRefSeg-R1 consists of three main components: a shared multi-scale vision encoder $E_v$, a change-aware multi-scale fusion module $\Phi$, and a language-guided mask decoder.
The two temporal images are first processed by the shared vision encoder to obtain multi-level visual features:
\begin{equation}
\begin{aligned}
\{\mathbf{F}^{t_1}_l\}_{l=1}^{L} &= E_v(\mathbf{I}^{t_1}), \qquad
\{\mathbf{F}^{t_2}_l\}_{l=1}^{L} = E_v(\mathbf{I}^{t_2}), \\
\mathbf{F}^{\Delta}_l &= \Phi_l(\mathbf{F}^{t_1}_l, \mathbf{F}^{t_2}_l), \qquad l=1,\dots,L .
\end{aligned}
\label{eq:overall_visual}
\end{equation}
Here, $\mathbf{F}^{t_1}_l$ and $\mathbf{F}^{t_2}_l$ denote the $l$-th level visual features of the earlier and later images, respectively, and $\mathbf{F}^{\Delta}_l$ denotes the corresponding change-aware fused feature.
In our implementation, the vision encoder produces four-scale features, namely $\texttt{res2}$, $\texttt{res3}$, $\texttt{res4}$, and $\texttt{res5}$.
Instead of concatenating the two RGB images at the input level, MTRefSeg-R1 performs feature-level temporal fusion at each scale.
This design preserves low-level spatial details while capturing high-level semantic changes.

For language reasoning, we reformulate the input prompt with explicit temporal order:
\begin{quote}
\small
\noindent
\texttt{``<image\_t1> is the earlier image, and <image\_t2> is the later image. Compare the two images and segment the <Referring Instruction>.''}
\end{quote}
This prompt informs the LVLM that the two images correspond to the same scene observed at different times.
The bi-temporal visual tokens and the textual instruction are processed by the multimodal language model $\mathcal{G}$.
Following segmentation-oriented LVLMs, we extract the hidden state of the generated $[\mathrm{SEG}]$ token and project it into the mask-query space:
\begin{equation}
\mathbf{q}_{\mathrm{seg}} = \mathcal{P}\big(\mathbf{h}_{[\mathrm{SEG}]}\big),
\label{eq:seg_query}
\end{equation}
where $\mathcal{P}(\cdot)$ denotes a learnable projector, $\mathbf{h}_{[\mathrm{SEG}]}$ is the hidden state of the $[\mathrm{SEG}]$ token, and $\mathbf{q}_{\mathrm{seg}}$ is the language-conditioned segmentation query.
The query $\mathbf{q}_{\mathrm{seg}}$ encodes which temporal change should be segmented, while the fused visual features $\{\mathbf{F}^{\Delta}_l\}_{l=1}^{L}$ provide dense evidence for locating the corresponding changed region.

\subsubsection{Change-Aware Multi-Scale Fusion}

The core of MTRefSeg-R1 is the \textbf{Change-Aware Fusion Block} $\Phi$, which explicitly models temporal differences between $\mathbf{I}^{t_1}$ and $\mathbf{I}^{t_2}$ at each feature level.
For the $l$-th level, the two temporal features are first projected into a unified feature space.
We then compute both signed and absolute temporal differences.
The signed difference preserves temporal direction from $t_1$ to $t_2$, which is useful for distinguishing appearance, disappearance, and state changes.
The absolute difference highlights the magnitude of local changes regardless of direction.
The fusion process is summarized as:
\begin{equation}
\begin{aligned}
\hat{\mathbf{F}}^{t_1}_l &= \psi^{t_1}_l(\mathbf{F}^{t_1}_l), \qquad
\hat{\mathbf{F}}^{t_2}_l = \psi^{t_2}_l(\mathbf{F}^{t_2}_l), \\
\mathbf{D}^{\mathrm{s}}_l &= \hat{\mathbf{F}}^{t_2}_l - \hat{\mathbf{F}}^{t_1}_l, \qquad
\mathbf{D}^{\mathrm{a}}_l = \left|\hat{\mathbf{F}}^{t_2}_l - \hat{\mathbf{F}}^{t_1}_l\right|, \\
\mathbf{C}_l &= \eta_l\big([\mathbf{D}^{\mathrm{s}}_l, \mathbf{D}^{\mathrm{a}}_l]\big), \\
\mathbf{U}_l &= \rho_l\big([\hat{\mathbf{F}}^{t_1}_l, \hat{\mathbf{F}}^{t_2}_l, \mathbf{D}^{\mathrm{a}}_l, \mathbf{C}_l]\big), \\
\mathbf{G}_l &= \sigma\Big(\omega_l\big(\mathrm{GAP}([\hat{\mathbf{F}}^{t_1}_l, \hat{\mathbf{F}}^{t_2}_l])\big)\Big), \\
\mathbf{F}^{\Delta}_l &= \delta\Big(\mathrm{Norm}\big(
\mathbf{U}_l + \mathbf{G}_l \odot \mathbf{C}_l
+ \frac{1}{2}(\hat{\mathbf{F}}^{t_1}_l+\hat{\mathbf{F}}^{t_2}_l)
\big)\Big).
\end{aligned}
\label{eq:change_aware_fusion}
\end{equation}
Here, $\psi^{t_1}_l(\cdot)$ and $\psi^{t_2}_l(\cdot)$ are timestamp-specific projection layers, $\mathbf{D}^{\mathrm{s}}_l$ and $\mathbf{D}^{\mathrm{a}}_l$ denote signed and absolute temporal differences, and $\eta_l(\cdot)$ encodes explicit change cues.
The function $\rho_l(\cdot)$ merges temporal semantics and change features into an intermediate representation $\mathbf{U}_l$.
The gate $\mathbf{G}_l$ is generated from globally pooled bi-temporal features and adaptively controls how much explicit change information should be injected into the fused representation.
Finally, the residual average term preserves stable semantic context shared by the two timestamps.

Compared with simple averaging, direct concatenation, or absolute-difference fusion, the proposed Change-Aware Fusion Block produces a more discriminative representation for language-specified temporal-change segmentation.

\subsubsection{Language-Guided Mask Decoding}

After multi-scale change-aware fusion, the fused features are fed into a pixel decoder $D_{\mathrm{pix}}$ to obtain high-resolution mask features and multi-scale decoder features.
The projected $[\mathrm{SEG}]$ embedding is then used as the language-conditioned query for the transformer mask decoder $D_{\mathrm{mask}}$:
\begin{equation}
\mathbf{S} =
D_{\mathrm{mask}}\big(D_{\mathrm{pix}}(\{\mathbf{F}^{\Delta}_l\}_{l=1}^{L}), \mathbf{q}_{\mathrm{seg}}\big),
\qquad
\mathbf{M} = \sigma(\mathbf{S}),
\label{eq:mask_decoding}
\end{equation}
where $\mathbf{S}$ denotes the predicted mask logit map and $\sigma(\cdot)$ is the sigmoid function.
In this process, the LVLM branch determines \emph{which} temporal change is referred to by the instruction, while the change-aware visual branch determines \emph{where} the corresponding changed region is located.
Therefore, MTRefSeg-R1 differs from conventional change detection models that segment all changed areas.
It performs language-conditioned change segmentation and only outputs the mask associated with the target described by the referring expression.

\subsubsection{Training Objective}

MTRefSeg-R1 is optimized with a joint objective consisting of language modeling and mask prediction losses:
\begin{equation}
\mathcal{L} =
\lambda_{\mathrm{llm}}\mathcal{L}_{\mathrm{llm}}
+
\lambda_{\mathrm{cls}}\mathcal{L}_{\mathrm{cls}}
+
\lambda_{\mathrm{mask}}\mathcal{L}_{\mathrm{mask}}
+
\lambda_{\mathrm{dice}}\mathcal{L}_{\mathrm{dice}}.
\label{eq:training_objective}
\end{equation}
Here, $\mathcal{L}_{\mathrm{llm}}$ supervises autoregressive language generation and encourages the model to generate the segmentation-related $[\mathrm{SEG}]$ token.
The term $\mathcal{L}_{\mathrm{cls}}$ supervises segmentation query classification, while $\mathcal{L}_{\mathrm{mask}}$ and $\mathcal{L}_{\mathrm{dice}}$ optimize pixel-level mask prediction.
The mask loss improves pixel-wise discrimination, and the Dice loss alleviates foreground-background imbalance, which is particularly important for small or sparse changed regions.

Overall, MTRefSeg-R1 preserves the reasoning ability of the base LVLM while introducing explicit multi-scale temporal-change modeling.
Through multi-temporal vision pretraining and referring multi-temporal fine-tuning, the model first learns general temporal-change perception and then acquires fine-grained language-conditioned change grounding.
As a result, MTRefSeg-R1 can understand temporally ordered image pairs, ground referring expressions, and generate accurate masks for language-specified changed regions.

%% file: table/1_datasetAnalysis_0.tex
\begin{figure*}[htbp]
    \centering
    \begin{subfigure}[b]{0.67\linewidth}
        \centering
        \includegraphics[width=\linewidth]{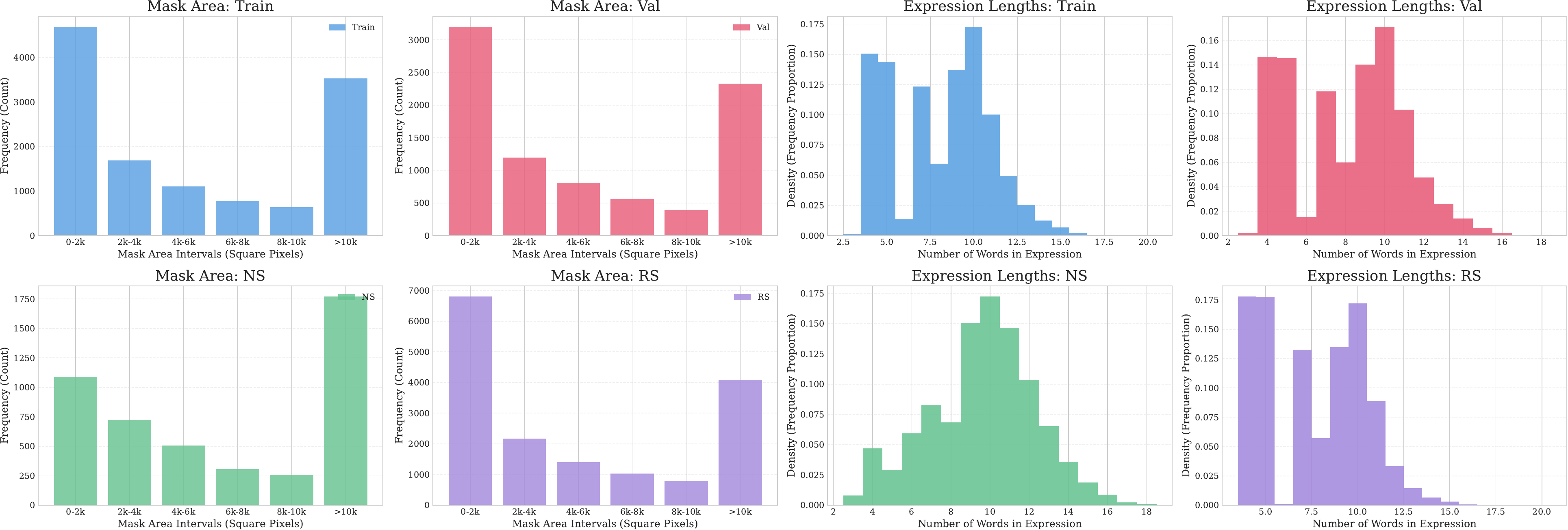}
    \end{subfigure}
    \hfill
    \begin{subfigure}[b]{0.30\linewidth}
        \centering
        \includegraphics[width=\linewidth]{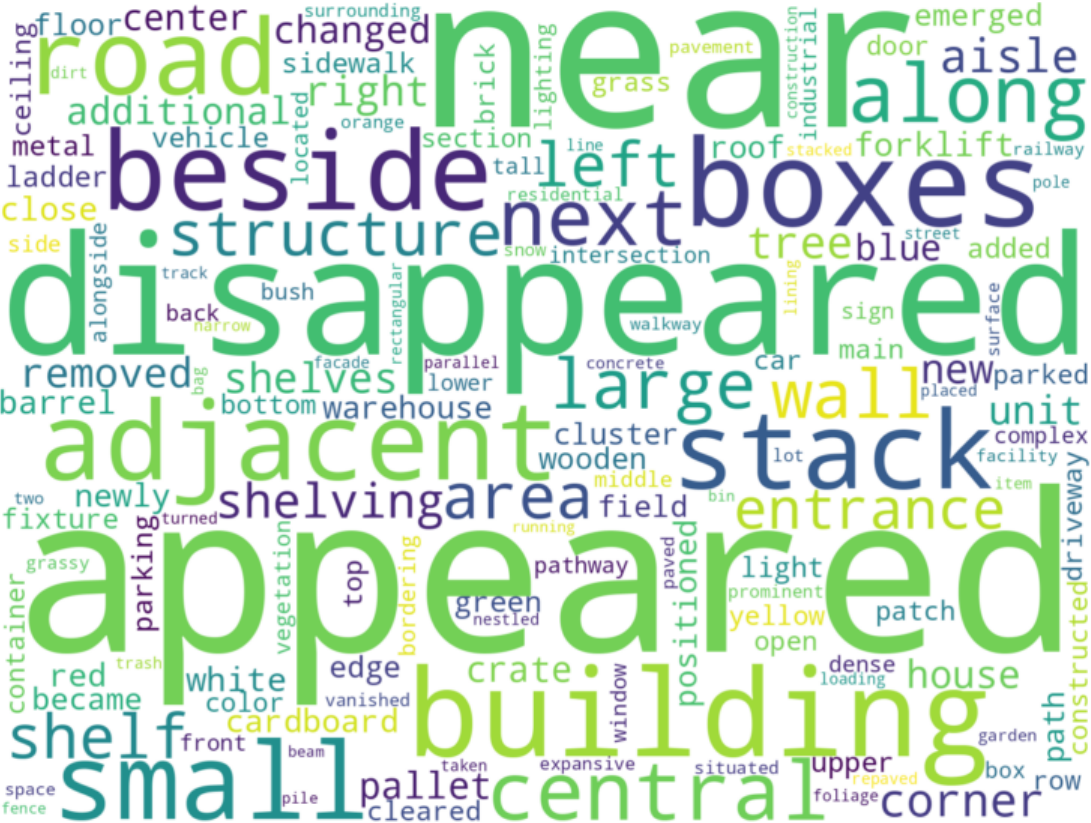}
    \end{subfigure}
    \caption{\textbf{Statistical analysis of the MTRefSeg-21K dataset.} 
    Left: Distribution of mask areas across the Train, Val, NS, and RS splits. 
    Middle: Distribution of referring expression lengths for the four splits, measured by the number of words per expression. 
    Right: Word cloud visualization of referring expressions, highlighting the most frequent words in the dataset.}
    \label{fig:three_images}
    \vspace{-1em}
\end{figure*}

%% file: table/1_datasetAnalysis_1.tex
\begin{table*}[htbp]
    \centering
        \caption{
        \textbf{Dataset statistics and multi-domain comparison.}
        Left: Detailed split distribution of MTRefSeg-21K.
        Right: Comparison with general-domain and remote-sensing referring segmentation datasets.
        MTRefSeg-21K features bi-temporal inputs, diverse resolutions, and multi-domain coverage across both normal-scene and remote-sensing domains.
        }
    \label{tab:combined_dataset_tables}
    \resizebox{\linewidth}{!}{
        \begin{minipage}[c]{0.4\linewidth} 
            \centering
            \setlength{\tabcolsep}{8pt}
            \renewcommand{\arraystretch}{1.55}
            \resizebox{\columnwidth}{!}{
            \begin{tabular}{lcc}
                \toprule
                \textbf{Split} & \textbf{Images} & \textbf{Expressions} \\ 
                \midrule
                \rowcolor{tabblue!5} 
                Train & 5646 & 12438 \\
                \rowcolor{tabblue!5} 
                Val  & 3875  & 8486  \\
                \midrule
                \rowcolor{tabpink!5} 
                 NS    & 1396 & 4654\\
                 \rowcolor{tabpink!5} 
                 RS    & 8125 & 16270  \\
                 \midrule
                 \rowcolor{gray!15} 
                 Total    & 9521 & 20924 \\
                \bottomrule
            \end{tabular}
            }
        \end{minipage}
        \hfill 
        \begin{minipage}[c]{0.70\linewidth}
            \centering
            \small
            \setlength{\tabcolsep}{4pt}
            \resizebox{\columnwidth}{!}{
            \begin{tabular}{lccccc}
                \toprule
                \textbf{Dataset} & \textbf{Task} & \textbf{Images} & \textbf{Expressions} & \textbf{Resolution} & \textbf{Bi-T} \\ 
                \midrule
                \multicolumn{6}{l}{\textit{General Domain}} \\
                gRefCOCO\cite{liu2023gres} & GRES & 19,994 & 278,232 & - & \xmark \\
                RefCOCOg\cite{yu2016modeling} & RES  & 26,711 & 85,474  & - & \xmark \\
                RefCOCO\cite{yu2016modeling}  & RES  & 19,994 & 142,209 & - & \xmark \\
                RefCOCO+\cite{yu2016modeling} & RES  & 19,992 & 141,564 & - & \xmark \\
                \midrule
                \multicolumn{6}{l}{\textit{Remote Sensing}} \\
                RefSegRS\cite{yuan2024rrsis} & R-RES & 4,420  & 4,420 & $512^2$ & \xmark \\
                RRSIS-D\cite{liu2024rotated}  & R-RES & 17,402 & 17,402 & $800^2$ & \xmark \\
                \midrule
                \rowcolor{gray!15} 
                \textbf{MTRefSeg-21K} & \textbf{MTRS} & \textbf{9521$\times$2} & \textbf{20924} & \textbf{224$^2\sim$2048$^2$} & \textbf{\cmark} \\ 
                \bottomrule
            \end{tabular}
            }
        \end{minipage}
    } 
    \vspace{-1em}
\end{table*}

%% file: sec/4_experiment.tex
\section{Experiments And Results}
\label{experiments_results}

\input{table/0_MainTable_0}
\input{table/0_MainTable_1}
\input{table/0_MainTable_trainval_0509}
\input{table/0_MainTable_ns_0509}
\input{table/0_MainTable_rs_0509}

\subsection{Implementation Details of the Methods}
For all compared methods, we first perform inference using their corresponding \emph{publicly available pretrained checkpoints}, and then fine-tune them on MTRefSeg-21K for downstream adaptation. Unless otherwise specified, the fixed hyperparameters follow the default settings in their official public implementations.
Following prior referring segmentation works, we report mIoU and cIoU for specialist models when applicable. 
For LVLM-based methods, we report oIoU under the same mask-level evaluation protocol. 
For compact presentation, we denote this column as cIoU/oIoU.
\begin{itemize}
    \item \textbf{Specialist Models.} All specialist baselines (CRIS, RefSegformer, LAVT, LGCE, RSRefSeg, RMSIN, FIANet) follow their original configurations, using \textbf{BERT} as the text encoder and either \textbf{Swin-B} or \textbf{ResNet-50/101} as the visual backbone. Following the same evaluation protocol, we fine-tune them on MTRefSeg-21K. Unless otherwise specified, the fixed hyperparameters follow the default settings in their official public implementations.
    \item \textbf{GeoPixel.} Initialized from the public \texttt{GeoPixel -7B-RES} checkpoint. We first evaluate the pretrained model and then fine-tune it on MTRefSeg-21K. Training uses 4 GPUs, \texttt{bf16}, LoRA, gradient checkpointing, and DeepSpeed ZeRO-2, with 3 epochs, learning rate \(3\times10^{-4}\), gradient accumulation steps 8, and maximum text length 4096.
    \item \textbf{GLaMM.} Initialized from \texttt{GLaMM-GranD-Pretr ained}, together with the public \texttt{SAM ViT-H} weights and \texttt{CLIP ViT-L/14} vision tower. We first evaluate the pretrained model and then fine-tune it on MTRefSeg-21K under the \textit{train$\rightarrow$val}, \textit{NS$\rightarrow$RS}, and \textit{RS$\rightarrow$NS} settings. Training uses DeepSpeed and LoRA, with 20 epochs and 500 steps per epoch.
    \item \textbf{GSVA-7B / GSVA-13B.} Initialized from the public \texttt{GSVA-7B} / \texttt{GSVA-13B} checkpoint, together with the public \texttt{SAM ViT-H} checkpoint and \texttt{CLIP ViT-L/14} vision tower. We first evaluate the pretrained model and then fine-tune it on MTRefSeg-21K under the three standard settings. Training uses DeepSpeed, \texttt{bf16}, and LoRA with \texttt{lora\_r=8}.
    \item \textbf{LISA-7B / LISA-13B.} Initialized from the public \texttt{LISA-7B-v1} and \texttt{LISA-13B-llama2-v1} checkpoints, respectively, together with the public \texttt{CLIP ViT-L/14} vision tower and \texttt{SAM ViT-H}. We first evaluate the pretrained models and then fine-tune them on MTRefSeg-21K. LISA-7B is trained for 20 epochs, while LISA-13B is trained for 20 epochs.
    \item \textbf{UniGeoSeg.} Initialized from its public pretrained checkpoint. We first conduct inference with the pretrained model and then fine-tune it on MTRefSeg-21K. The provided setting uses image size 512 and evaluation batch size 4. Other fixed settings follow the official code.
    \item \textbf{SegEarth-R1.} Initialized from the public \texttt{SegEarth -R1-RRSIS-D} checkpoint. We first evaluate the pretrained model and then fine-tune it on MTRefSeg-21K. The provided setting uses image size 1024 and batch size 16. Other fixed settings follow the official code.
    \item \textbf{UniChange.} UniChange is included as a recent change-oriented LVLM baseline. It uses CLIP-L as the visual encoder and Vicuna-7B as the language backbone. Since UniChange is designed for change-aware visual understanding, we include it to compare MTRS-specific adaptation with existing change-oriented LVLMs.
    \item \textbf{MTRefSeg-R1.} For MTRefSeg-R1, Stage 1 is trained on approximately 20K vision-only bi-temporal samples, and Stage 2 is fully fine-tuned on MTRefSeg-21K. 
    We use Swin-B as the visual encoder and Phi-1.5 as the language backbone. 
\end{itemize}

\subsection{Comparisons with Existing LVLMs}
\label{sec:comparison_lvlms}

\subsubsection{Overall Mean Performance}

Table~\ref{tab:changeref_results_mean} reports the average performance across the Train$\rightarrow$Val, NS$\rightarrow$NS, and RS$\rightarrow$RS settings. 
The results reveal a clear gap between direct LVLM inference and task-specific adaptation. 
When existing LVLMs are directly evaluated without fine-tuning, their performance remains very limited, with most models achieving only around 2--18 mIoU. 
This indicates that single-temporal vision-language pretraining alone is insufficient for temporal correspondence reasoning and language-guided change localization.

After fine-tuning on MTRefSeg-21K, LVLM baselines improve substantially, demonstrating that task-specific supervision is necessary for adapting general LVLMs to MTRS. 
Nevertheless, these fine-tuned LVLMs still lag behind our method. 
MTRefSeg-R1 achieves the best mean mIoU of 65.68 and the best mean Pr@50 of 71.65, outperforming the strongest fine-tuned LVLM baseline, SegEarth-R1, by 6.08 mIoU and 6.67 Pr@50. 
Although several specialist segmentation methods remain competitive in cIoU/oIoU, our method achieves the strongest overall balance between segmentation accuracy and high-quality referred-change localization among LVLM-based approaches.

\subsubsection{Detailed Evaluation under the Train$\rightarrow$Val Setting}

To further analyze model behavior, Table~\ref{tab:train_to_val} reports detailed results under the Train$\rightarrow$Val setting, including mIoU, cIoU/oIoU, and precision-at-threshold metrics from Pr@50 to Pr@90. 
This setting evaluates the general performance of models when training and validation samples are drawn from the overall MTRefSeg-21K distribution.

Compared with direct LVLM inference, fine-tuning brings large improvements across all metrics, confirming that MTRS requires explicit adaptation to bi-temporal visual inputs and referring-change instructions. 
Our method achieves the best mIoU of 68.24 and obtains the highest precision scores from Pr@50 to Pr@90. 
In particular, the consistent advantage under stricter thresholds such as Pr@80 and Pr@90 suggests that MTRefSeg-R1 not only detects approximate changed regions, but also generates more accurate masks with better boundary quality and stronger alignment to the referred temporal targets.

\subsubsection{Benchmarking on the Normal-Scene Domain}

We further split the benchmark into NS and RS domains to provide domain-specific evaluation for different research communities. 
Table~\ref{tab:ns_to_ns} reports the NS$\rightarrow$NS setting, where models are trained and evaluated within the natural-scene domain. 
This setting focuses on normal-scene changes, where objects often exhibit diverse appearances, occlusions, illumination variations, and complex interactions with surrounding environments.

In this domain, several specialist methods and fine-tuned LVLM baselines achieve competitive mIoU scores, showing that normal-scene referring change segmentation benefits from strong object-level priors learned from general visual data. 
Our method still achieves strong performance, especially on high-overlap precision metrics. 
MTRefSeg-R1 obtains the best Pr@70 and remains highly competitive at Pr@60 and Pr@80, indicating that it can accurately localize language-specified temporal changes even when the scene contains cluttered backgrounds, shadows, or subtle object-level variations.

\subsubsection{Benchmarking on the Remote-Sensing Domain}

Table~\ref{tab:rs_to_rs} reports the RS$\rightarrow$RS setting, where models are trained and evaluated within the remote-sensing domain. 
This setting is particularly important for remote-sensing researchers, since aerial-view images contain dense layouts, small objects, large scale variations, and visually similar structures. 
These properties make language-guided temporal segmentation more challenging than conventional single-image referring segmentation.

Existing LVLMs without fine-tuning perform poorly in the RS domain, suggesting that general single-temporal LVLMs lack the ability to directly interpret remote-sensing temporal changes. 
Fine-tuning improves their performance, but most LVLM baselines still struggle to match specialist methods and our proposed model. 
MTRefSeg-R1 achieves the best mIoU of 68.92 and the best precision scores from Pr@50 to Pr@90, demonstrating its strong ability to capture building-level and region-level changes in aerial scenes. 
These results show that the proposed change-aware temporal fusion and two-stage training strategy are especially beneficial for remote-sensing MTRS, where accurate cross-temporal comparison is critical.

Overall, the results demonstrate three key observations. 
First, existing LVLMs cannot be directly transferred to MTRS without task-specific training, revealing the limitation of single-temporal vision-language pretraining. 
Second, fine-tuning on MTRefSeg-21K substantially improves LVLM performance, validating the importance of the proposed benchmark. 
Third, by separately reporting overall, NS-domain, and RS-domain results, MTRefSeg-21K provides a unified yet domain-aware benchmark for both general-scene and remote-sensing communities. 
MTRefSeg-R1 consistently achieves strong performance across these settings, showing its effectiveness for language-guided multi-temporal change understanding.

\subsection{Ablation Studies}
\label{sec:ablation}

We analyze the contribution of temporal fusion, pretraining strategy, and parameter-efficient tuning for bi-temporal referring change segmentation. Unless otherwise stated, all results are reported on the validation set. These ablations are designed to answer three key questions: how to effectively fuse bi-temporal visual features, whether explicit change-oriented pretraining is beneficial, and whether parameter-efficient tuning is sufficient for fine-grained multi-temporal referring segmentation.

\paragraph{Effect of temporal fusion.}
Table~\ref{tab:fusion_ablation} compares different temporal fusion strategies. In Stage-1 visual pretraining, all explicit bi-temporal fusion schemes outperform the naive \texttt{avg} baseline, showing that modeling temporal differences is essential for learning change-aware visual representations. The best Stage-1 performance is achieved by \texttt{abs\_diff}, which directly emphasizes changed regions through feature differencing. However, its advantage over \texttt{change\_aware} is marginal, while pure differencing becomes less effective after language-guided multimodal fine-tuning.

In Stage-2, the final \texttt{change\_aware} configuration achieves 68.24 mIoU and 67.82 oIoU. This shows that \texttt{change\_aware} provides a balanced fusion mechanism: it preserves shared semantic context, captures explicit temporal differences, and supports language-conditioned localization. Compared with simple averaging or pure absolute differencing, the proposed design is more suitable for MTRS, where the model must not only detect visual changes but also identify the specific changed region described by the referring expression.

\begin{table}[t]
\centering
\small
\caption{Temporal fusion ablation on Stage-1 visual pretraining and Stage-2 multimodal fine-tuning.}
\label{tab:fusion_ablation}
\setlength{\tabcolsep}{4pt}
\resizebox{\linewidth}{!}{
\begin{tabular}{lcccc}
\toprule
\multirow{2}{*}{Fusion Type} 
& \multicolumn{2}{c}{Stage-1 Visual Pretraining} 
& \multicolumn{2}{c}{Stage-2 Multimodal Fine-tuning} \\
\cmidrule(lr){2-3} \cmidrule(lr){4-5}
& mIoU & oIoU & mIoU & oIoU \\
\midrule
\texttt{avg}          & 81.54 & 86.69 & 66.47 & 66.39 \\
\texttt{abs\_diff}    & \textbf{83.04} & \textbf{88.21} & 66.14 & 66.71 \\
\texttt{concat\_conv} & 82.73 & 87.75 & \textbf{68.90} & \textbf{68.45} \\
\texttt{change\_aware} & 82.60 & 87.98 & 68.24 & 67.82 \\
\bottomrule
\end{tabular}
}
\end{table}

\paragraph{Effect of pretraining and initialization.}
Table~\ref{tab:pretrain_ablation} studies the effect of different pretraining and initialization strategies. Compared with end-to-end multimodal pretraining, the dedicated Stage-1 visual change pretraining achieves substantially better pretraining performance, improving mIoU from 79.02 to 82.65 and oIoU from 85.06 to 87.92. This suggests that learning temporal-change perception in a vision-focused manner provides a cleaner and more effective supervisory signal before introducing language instructions.

The benefit of Stage-1 pretraining is also preserved during Stage-2 multimodal fine-tuning. Initializing Stage-2 from the Stage-1 pretrained model outperforms both end-to-end MLLM pretrained initialization and training without pretraining. This demonstrates that generic temporal-change priors are important for MTRS. Instead of forcing the model to learn temporal correspondence, language grounding, and mask prediction simultaneously from scratch, the two-stage recipe decomposes the problem into change-aware visual representation learning followed by fine-grained language-guided segmentation.

\begin{table}[t]
\centering
\small
\caption{Ablation on pretraining and initialization strategy.}
\label{tab:pretrain_ablation}
\setlength{\tabcolsep}{4pt}
\resizebox{\linewidth}{!}{
\begin{tabular}{lcc}
\toprule
Setting & mIoU & oIoU \\
\midrule
\multicolumn{3}{l}{\textit{Stage-1 pretraining strategy}} \\
Stage-1 visual pretraining & \textbf{82.65} & \textbf{87.92} \\
End-to-end MLLM pretraining & 79.02 & 85.06 \\
\midrule
\multicolumn{3}{l}{\textit{Stage-2 initialization strategy}} \\
Stage-1 pretrained initialization & \textbf{68.24} & \textbf{67.82} \\
End-to-end MLLM pretrained initialization & 67.45 & 67.21 \\
No pretraining & 67.12 & 67.29 \\
\bottomrule
\end{tabular}
}
\end{table}

\paragraph{Effect of fine-tuning strategy.}
Table~\ref{tab:finetuning_ablation} compares parameter-efficient LoRA fine-tuning with full fine-tuning for MTRS. 
Increasing the LoRA rank from 8 to 64 consistently improves performance, raising mIoU from 56.92 to 64.45 and oIoU from 57.84 to 63.84, indicating that MTRS requires sufficient adaptation capacity for temporal reasoning and mask prediction. 
However, full fine-tuning further improves the results to 68.24 mIoU and 67.82 oIoU, outperforming the strongest LoRA setting by 3.79 mIoU and 3.98 oIoU. 
This shows that low-rank adaptation is helpful but still insufficient for fully adapting the visual encoder, temporal fusion module, multimodal connector, and mask decoder. 
Therefore, we adopt full multimodal fine-tuning as the final strategy for MTRefSeg-R1.

\begin{table}[t]
\centering
\small
\caption{Ablation on fine-tuning strategy.}
\label{tab:finetuning_ablation}
\setlength{\tabcolsep}{5pt}
\begin{tabular}{lcc}
\toprule
Fine-tuning Strategy & mIoU & oIoU \\
\midrule
LoRA fine-tuning, $r=8$  & 56.92 & 57.84 \\
LoRA fine-tuning, $r=16$ & 60.19 & 59.92 \\
LoRA fine-tuning, $r=32$ & 62.30 & 61.51 \\
LoRA fine-tuning, $r=64$ & 64.45 & 63.84 \\
\rowcolor{yellow!20}
Full fine-tuning, Ours & \textbf{68.24} & \textbf{67.82} \\
\bottomrule
\end{tabular}
\end{table}

\begin{figure}[t]
\centering
\includegraphics[width=\linewidth]{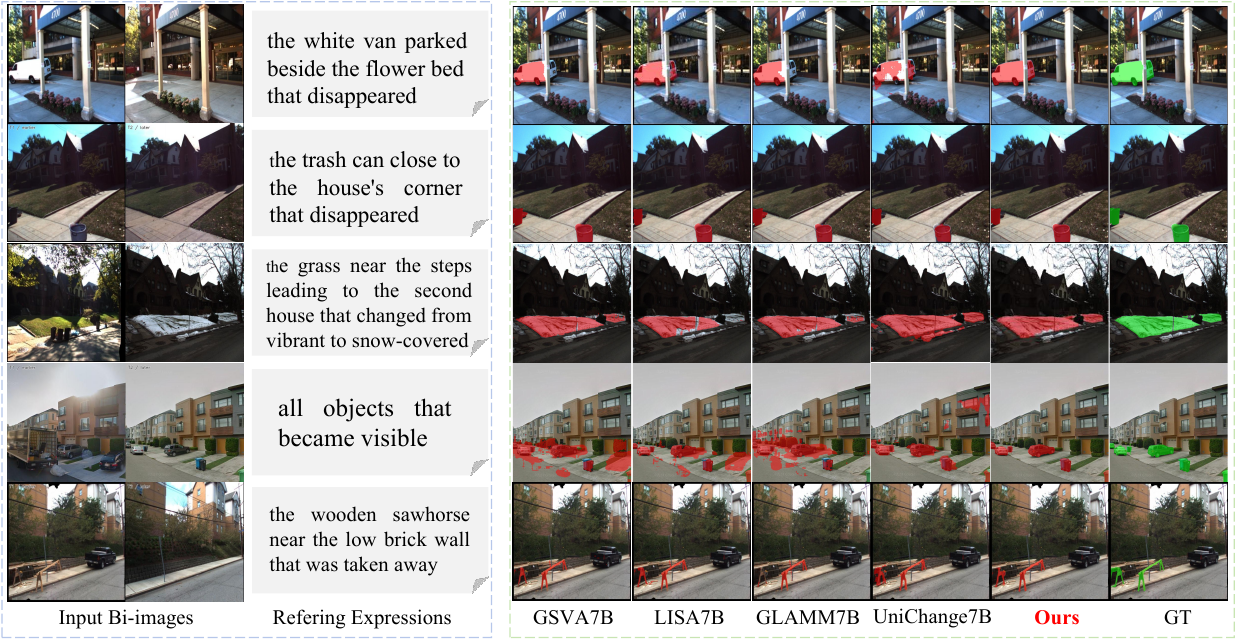}
\caption{
\textbf{Qualitative comparisons on the NS domain.}
We compare our method with representative LVLM baselines under normal-scene multi-temporal referring segmentation.
The examples cover object disappearance, appearance, and state changes.
Compared with existing LVLMs, our method produces more complete and spatially accurate masks that better match the language-specified temporal changes.
}
\label{fig:pic_vis_NS}
\vspace{-1em}
\end{figure}

\begin{figure}[t]
\centering
\includegraphics[width=\linewidth]{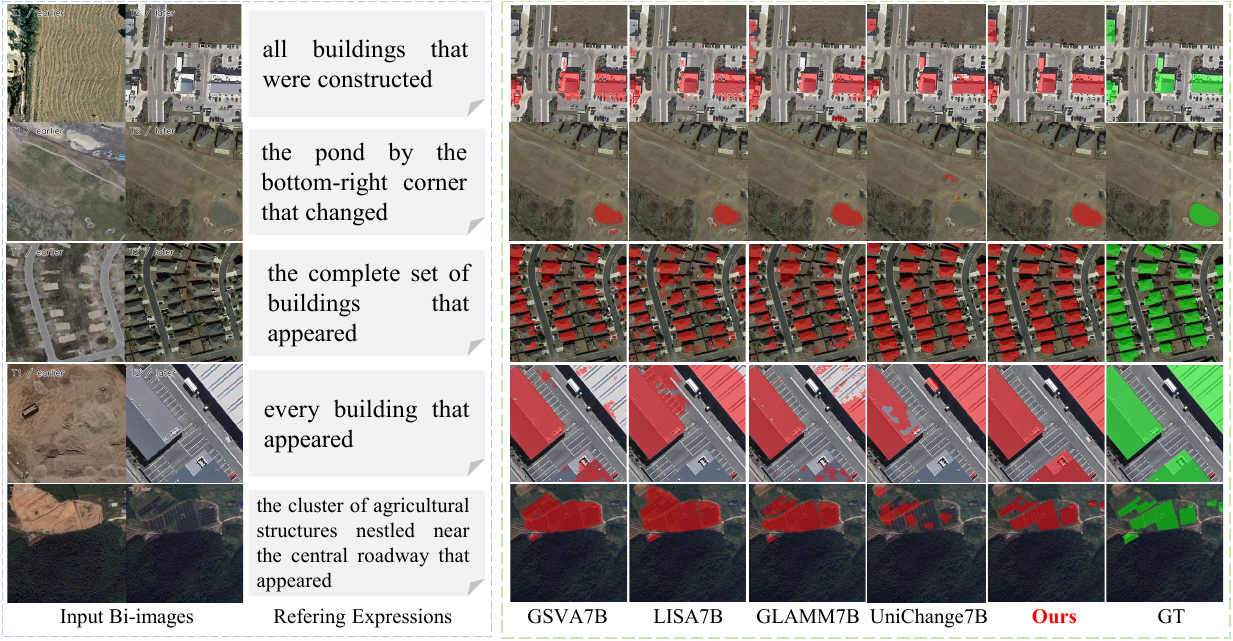}
\caption{
\textbf{Qualitative comparisons on the RS domain.}
We compare our method with representative LVLM baselines on remote-sensing multi-temporal referring segmentation.
The results show that our method can better capture building-level and region-level temporal changes under aerial-view scenes, producing masks that are more consistent with the ground truth.
}
\label{fig:pic_vis_RS}
\vspace{-1em}
\end{figure}

\subsection{Visualization}

To further demonstrate the effectiveness of MTRefSeg-R1, we provide qualitative comparisons on both normal-scene and remote-sensing domains. As shown in Fig.~\ref{fig:pic_vis_NS}, existing LVLM-based segmentation models can roughly respond to the referring expressions, but they often suffer from incomplete masks, inaccurate boundaries, or confusion between language-relevant changes and irrelevant temporal differences. For example, when the expression refers to a disappeared object or a newly visible object, baseline models may segment only part of the target or include surrounding unchanged regions. In contrast, our method produces masks that are more spatially complete and better aligned with the specified temporal change.

Fig.~\ref{fig:pic_vis_RS} shows qualitative results on the remote-sensing domain. Compared with normal-scene images, remote-sensing scenes contain smaller objects, denser layouts, and more ambiguous boundaries, making language-guided temporal segmentation more challenging. Existing models tend to over-segment adjacent buildings or miss small changed structures. Our method achieves more accurate localization of building-level and region-level changes, demonstrating stronger temporal correspondence reasoning and better adaptation to aerial-view scenes.

\begin{figure}[t]
    \centering
    \includegraphics[width=\linewidth]{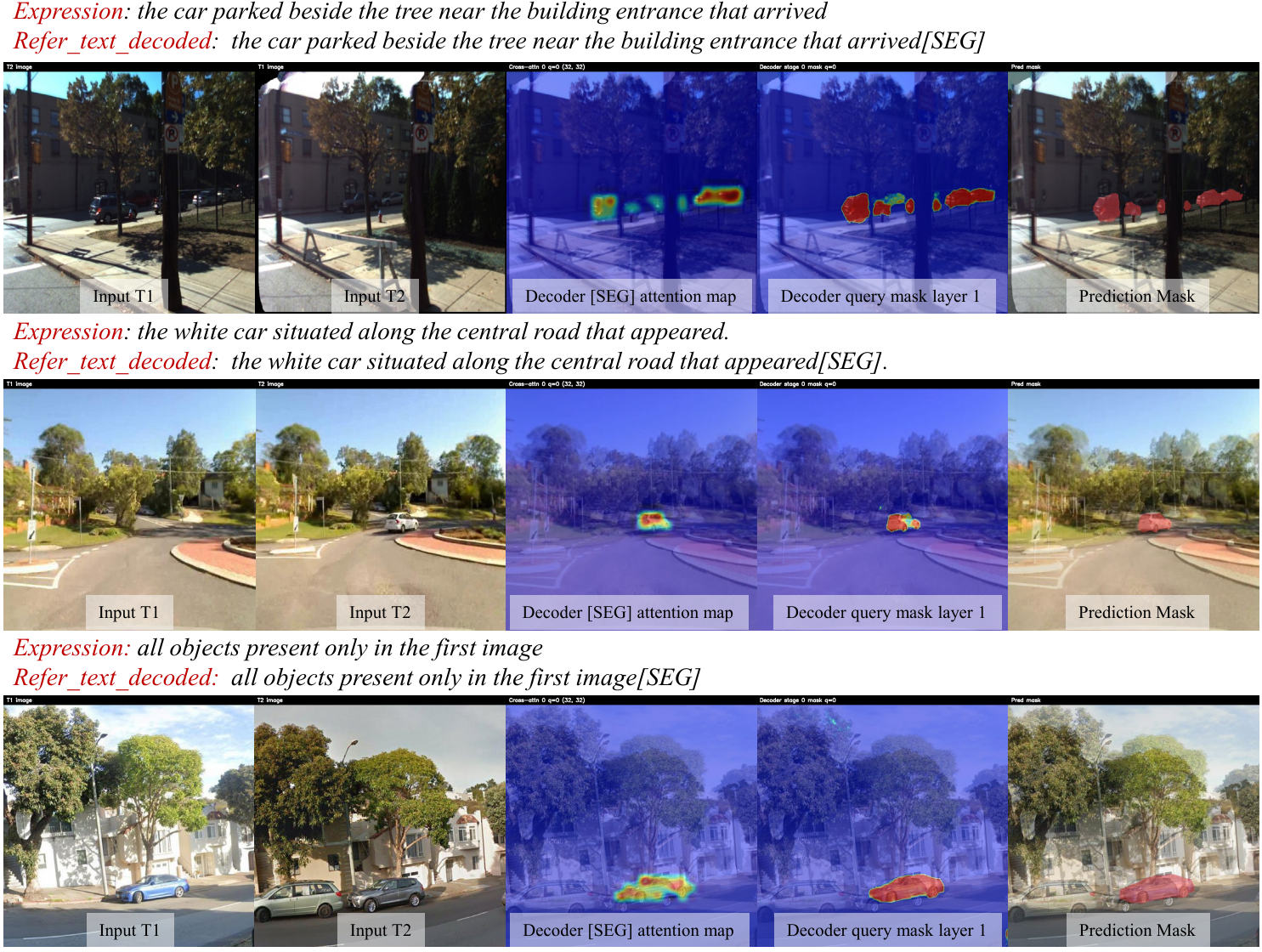}
    \caption{
    \textbf{Visualization of language-guided temporal attention and mask decoding.}
    The decoder \texttt{[SEG]} attention maps concentrate on the regions described by the referring expressions, and the intermediate query masks progressively localize the target changed objects before producing the final prediction.
    }
    \label{fig:pic_vis_attn}
    \vspace{-0.8em}
\end{figure}

We further visualize the internal decoding behavior in Fig.~\ref{fig:pic_vis_attn}. The decoder \texttt{[SEG]} attention maps concentrate on the regions described by the referring expressions, while the intermediate query masks progressively highlight the target changed objects. These results indicate that the proposed model can align language instructions with bi-temporal visual differences and convert the language-conditioned \texttt{[SEG]} representation into accurate pixel-level masks. This provides qualitative evidence that MTRefSeg-R1 performs language-guided change reasoning rather than simply detecting all changed regions.

%% file: table/0_MainTable_1.tex
\begin{table}[htbp]
    \centering
    \scriptsize
\renewcommand{\arraystretch}{0.88}
    \caption{\textbf{Quantitative comparison results on MTRefSeg-21K in terms of mean performance.}
    We report the mean performance across the Train$\rightarrow$Val, NS$\rightarrow$NS, and RS$\rightarrow$RS settings, including mIoU, cIoU/oIoU, and Pr@50.
    Best results in each column are highlighted in \textbf{bold}.}
    \label{tab:changeref_results_mean}
    \setlength{\tabcolsep}{6pt}
    \large
    \begin{adjustbox}{max width=\linewidth}
    \begin{tabular}{l c c c c}
        \toprule
        \multirow{2}{*}{Model}
        & \multirow{2}{*}{Vision Encoder}
        & \multicolumn{3}{c}{Mean} \\
        \cmidrule(lr){3-5}
        &
        & mIoU & cIoU/oIoU & Pr@50 \\
        \midrule
        \rowcolor{gray!15}
        \multicolumn{5}{l}{\textit{\textbf{Specialist Methods, Trained on MTRefSeg-21K}}} \\
        CRIS-50\cite{wang2022cris}\pub{CVPR2022}             & ResNet-50  & 48.01 & --    & 51.23 \\
        CRIS-101\cite{wang2022cris}\pub{CVPR2022}            & ResNet-101 & 48.71 & --    & 51.94 \\
        RefSegformer\cite{wu2024toward}\pub{TIP2024}         & Swin-B     & 61.37 & 64.87 & 68.08 \\
        LAVT\cite{yang2022lavt}\pub{CVPR2022}                & Swin-B     & 62.60 & 66.02 & 68.94 \\
        LGCE\cite{yuan2024rrsis}\pub{TGRS2024}               & Swin-B     & 63.27 & \uline{66.22} & 69.87 \\
        RSRefSeg\cite{chen2025rsrefseg1}\pub{IGARSS2025}     & Swin-B     & 59.08 & 57.59 & 64.02 \\
        RMSIN\cite{liu2024rotated}\pub{CVPR2024}             & Swin-B     & 64.35 & 66.07 & \uline{71.50} \\
        FIANet\cite{lei2024exploring}\pub{TGRS2024}          & Swin-B     & \uline{64.73} & \textbf{67.87} & 70.07 \\
        \midrule

        \rowcolor{gray!15}
        \multicolumn{5}{l}{\textit{\textbf{LVLMs with Single-Temporal Pretrain, Without Fine-tuning}}} \\
        LISA-7B\cite{lai2024lisa}\pub{CVPR2024}              & CLIP-L     &  5.54 &  6.97 &  1.52 \\
        LISA-13B\cite{lai2024lisa}\pub{CVPR2024}             & CLIP-L     &  2.46 &  2.58 &  0.67 \\
        GSVA-7B\cite{xia2024gsva}\pub{CVPR2024}              & CLIP-L     & 14.35 & 14.88 &  7.46 \\
        GSVA-13B\cite{xia2024gsva}\pub{CVPR2024}             & CLIP-L     & 17.77 & 18.17 & 12.03 \\
        GLAMM-7B\cite{rasheed2024glamm}\pub{CVPR2024}        & CLIP-L     & 11.51 &  9.63 &  8.61 \\
        GeoPixel-7B\cite{shabbir2025geopixel}\pub{ICML2025}  & CLIP-L     & 11.23 & 12.44 &  5.68 \\
        UniGeoSeg\cite{ni2025unigeoseg}\pub{CVPR2026}        & Swin-B     & 14.31 &  8.22 & 12.45 \\
        SegEarth-R1\cite{li2025segearth}\pub{Arkiv2026}       & Swin-B     &  6.92 &  4.57 &  4.56 \\
        UniChange\cite{zhang2025unichange}\pub{CVPR2026}     & CLIP-L     & --    & --    & --    \\
        \midrule

        \rowcolor{gray!15}
        \multicolumn{5}{l}{\textit{\textbf{LVLMs with Single-Temporal Pretrain,  Fine-tuned on MTRefSeg-21K}}} \\
        LISA-7B\cite{lai2024lisa}\pub{CVPR2024}              & CLIP-L     & 52.40 & 50.85 & 55.15 \\
        LISA-13B\cite{lai2024lisa}\pub{CVPR2024}             & CLIP-L     & 54.09 & 50.84 & 57.16 \\
        GSVA-7B\cite{xia2024gsva}\pub{CVPR2024}              & CLIP-L     & 56.62 & 52.56 & 61.07 \\
        GSVA-13B\cite{xia2024gsva}\pub{CVPR2024}             & CLIP-L     & 55.71 & 53.11 & 59.42 \\
        GLAMM-7B\cite{rasheed2024glamm}\pub{CVPR2024}        & CLIP-L     & 56.37 & 51.92 & 60.34 \\
        GeoPixel-7B\cite{shabbir2025geopixel}\pub{ICML2025}  & CLIP-L     & 16.40 & 16.22 &  9.86 \\
        SegEarth-R1\cite{li2025segearth}\pub{Arkiv2026}       & Swin-B     & 59.60 & 59.99 & 64.98 \\
        UniChange\cite{zhang2025unichange}\pub{CVPR2026}     & CLIP-L     & 51.24 & 45.30 & 53.88 \\
        \rowcolor{yellow!20}
        Ours                                                   & Swin-B     & \textbf{65.68} & 65.17 & \textbf{71.65} \\
        \bottomrule
    \end{tabular}
    \end{adjustbox}
    \vspace{-1em}
\end{table}

%% file: table/0_MainTable_trainval_0509.tex
\begin{table*}[t]
\centering
\scriptsize
\renewcommand{\arraystretch}{0.88}
\caption{
Comparison of specialist referring segmentation methods and large vision-language models under the \textbf{train$\rightarrow$val} setting.
The table reports \textit{mIoU}, \textit{cIoU/oIoU}, and precision-at-threshold metrics (\textit{Pr@50--Pr@90}).
Methods are divided into three groups:
(1) \textit{Specialist Methods}, including conventional referring segmentation models trained with dedicated vision-language backbones;
(2) \textit{LVLMs with Single-Temporal Pretrain, Without Fine-tuning}, where LVLMs are evaluated without fine-tuning on MTRefSeg-21K;
and (3) \textit{LVLMs with Single-Temporal Pretrain, fine-tuned on MTRefSeg-21K}, where LVLMs are adapted on the MTRefSeg-21K dataset.
}
\label{tab:train_to_val}
\resizebox{\textwidth}{!}{
\begin{tabular}{l l l c c c c c c c}
\toprule
\multirow{2}{*}{Model} & \multirow{2}{*}{Vision Encoder} & \multirow{2}{*}{Text Encoder} & \multicolumn{7}{c}{Train $\rightarrow$ Val}  \\
\cmidrule(lr){4-10}
 &  &  & mIoU & cIoU/oIoU & Pr@50 & Pr@60 & Pr@70 & Pr@80 & Pr@90 \\
\midrule

\rowcolor{gray!15}
\multicolumn{10}{l}{\textit{\textbf{Specialist Methods, Trained on MTRefSeg-21K}}} \\
CRIS-50\cite{wang2022cris}\pub{CVPR2022}          & ResNet-50  & ResNet-50             & 53.46 & -     & 58.37 & 51.00 & 41.63 & 25.01 & 13.56 \\
CRIS-101\cite{wang2022cris}\pub{CVPR2022}         & ResNet-101 & ResNet-101            & 53.93 & -     & 58.64 & 52.00 & 44.01 & 27.93 & 14.39 \\
RefSegformer\cite{wu2024toward}\pub{TIP2024}      & Swin-B     & BERT                  & 64.07 & 66.67 & 71.05 & 65.21 & 57.77 & 46.75 & 23.80 \\
LAVT\cite{yang2022lavt}\pub{CVPR2022}             & Swin-B     & BERT                  & 64.76 & 67.98 & 70.87 & 64.93 & 57.81 & 47.65 & 24.36 \\
LGCE\cite{yuan2024rrsis}\pub{TGRS2024}            & Swin-B     & BERT                  & 66.25 & \textbf{68.88} & 72.60 & 67.28 & \uline{60.38} & \uline{49.82} & 25.80 \\
RSRefSeg\cite{chen2025rsrefseg1}\pub{IGARSS2025}  & Swin-B     & BERT                  & 64.99 & 60.55 & 70.99 & 64.90 & 57.78 & 48.63 & \uline{26.87} \\
RMSIN\cite{liu2024rotated}\pub{CVPR2024}          & Swin-B     & BERT                  & \uline{66.51} & \uline{68.18} & \uline{73.31} & \uline{67.77} & 59.97 & 49.81 & 24.53 \\
FIANet\cite{lei2024exploring}\pub{TGRS2024}       & Swin-B     & BERT                  & 65.82 & 67.03 & 72.27 & 66.91 & 59.85 & 49.39 & 24.22 \\

\midrule

\rowcolor{gray!15}
\multicolumn{10}{l}{\textit{\textbf{LVLMs with Single-Temporal Pretrain, Without Fine-tuning}}} \\
LISA-7B\cite{lai2024lisa}\pub{CVPR2024}           & CLIP-L     & Vicuna-7B             &  5.84 &  7.32 &  1.68 &  1.01 &  0.43 &  0.14 &  0.05 \\
LISA-13B\cite{lai2024lisa}\pub{CVPR2024}          & CLIP-L     & Vicuna-13B            &  2.46 &  2.62 &  0.34 &  0.24 &  0.24 &  0.24 &  0.24 \\
GSVA-7B\cite{xia2024gsva}\pub{CVPR2024}           & CLIP-L     & Vicuna-7B             & 13.53 & 14.02 &  5.81 &  3.55 &  2.74 &  1.49 &  0.53 \\
GSVA-13B\cite{xia2024gsva}\pub{CVPR2024}          & CLIP-L     & Vicuna-13B            & 16.25 & 17.14 & 10.18 &  7.06 &  4.08 &  2.16 &  0.53 \\
GLAMM-7B\cite{rasheed2024glamm}\pub{CVPR2024}     & CLIP-L     & Vicuna-7B             & 12.67 & 11.10 & 10.13 &  7.64 &  5.96 &  3.99 &  2.69 \\
GeoPixel-7B\cite{shabbir2025geopixel}\pub{ICML2025} & CLIP-L  & InternLM-XComposer2.5 & 10.80 & 12.35 &  5.46 &  3.62 &  2.20 &  1.46 &  0.86 \\
UniGeoSeg\cite{ni2025unigeoseg}\pub{CVPR2026}     & Swin-B     & Phi-1.5               & 14.20 &  7.32 & 12.40 & 10.96 &  9.56 &  8.21 &  6.62 \\
SegEarth-R1\cite{li2025segearth}\pub{Arkiv2026}    & Swin-B     & Phi-1.5               &  7.26 &  4.42 &  5.08 &  4.36 &  3.63 &  3.10 &  2.67 \\

\midrule

\rowcolor{gray!15}
\multicolumn{10}{l}{\textit{\textbf{LVLMs with Single-Temporal Pretraining, Fine-tuned on MTRefSeg-21K}}} \\
LISA-7B\cite{lai2024lisa}\pub{CVPR2024}           & CLIP-L     & Vicuna-7B             & 51.42 & 49.78 & 54.51 & 44.28 & 35.01 & 26.46 & 17.48 \\
LISA-13B\cite{lai2024lisa}\pub{CVPR2024}          & CLIP-L     & Vicuna-13B            & 53.00 & 51.16 & 56.53 & 45.92 & 36.89 & 28.24 & 19.88 \\
GSVA-7B\cite{xia2024gsva}\pub{CVPR2024}           & CLIP-L     & Vicuna-7B             & 54.56 & 50.43 & 58.26 & 49.33 & 38.71 & 30.02 & 21.04 \\
GSVA-13B\cite{xia2024gsva}\pub{CVPR2024}          & CLIP-L     & Vicuna-13B            & 54.81 & 52.81 & 58.89 & 49.14 & 39.34 & 30.55 & 21.37 \\
GLAMM-7B\cite{rasheed2024glamm}\pub{CVPR2024}     & CLIP-L     & Vicuna-7B             & 54.68 & 50.40 & 58.36 & 48.17 & 38.81 & 30.21 & 19.60 \\
GeoPixel-7B\cite{shabbir2025geopixel}\pub{ICML2025} & CLIP-L  & InternLM-XComposer2.5 & 14.06 & 13.71 &  7.79 &  6.02 &  4.58 &  3.62 &  3.03 \\
SegEarth-R1\cite{li2025segearth}\pub{Arkiv2026}    & Swin-B     & Phi-1.5               & 61.46 & 62.55 & 66.98 & 62.11 & 55.50 & 44.64 & 24.92 \\
UniChange\cite{zhang2025unichange}\pub{CVPR2026}  & CLIP-L     & Vicuna-7B             & 54.42 & 49.44 & 57.88 & 50.91 & 42.84 & 31.84 & 16.38 \\
\rowcolor{yellow!20}
Ours                                    & Swin-B     & Phi-1.5              & \textbf{68.24} & 67.82 & \textbf{73.90} & \textbf{69.52} & \textbf{63.51} & \textbf{54.81} & \textbf{32.39} \\
\bottomrule
\end{tabular}
}
\end{table*}

%% file: table/0_MainTable_ns_0509.tex
\begin{table*}[t]
\centering
\scriptsize
\renewcommand{\arraystretch}{0.88}
\caption{
Comparison of different models under the \textbf{NS$\rightarrow$NS} validation setting, where models are trained and evaluated within the natural-scene domain.
This setting evaluates in-domain performance on natural-scene change referring segmentation, where the training and validation data share similar visual appearance, object scale, and scene layout distributions.
}
\label{tab:ns_to_ns}
\resizebox{\textwidth}{!}{
\begin{tabular}{l l l c c c c c c c}
\toprule
\multirow{2}{*}{Model} & \multirow{2}{*}{Vision Encoder} & \multirow{2}{*}{Text Encoder} & \multicolumn{7}{c}{NS Train $\rightarrow$ NS Val}  \\
\cmidrule(lr){4-10}
 &  &  & mIoU & cIoU/oIoU & Pr@50 & Pr@60 & Pr@70 & Pr@80 & Pr@90 \\
\midrule

\rowcolor{gray!15}
\multicolumn{10}{l}{\textit{\textbf{Specialist Methods, Trained on MTRefSeg-21K}}} \\
CRIS-50\cite{wang2022cris}\pub{CVPR2022}          & ResNet-50  & ResNet-50  & 35.24 & -     & 35.04 & 25.81 & 14.99 &  6.72 &  1.30 \\
CRIS-101\cite{wang2022cris}\pub{CVPR2022}         & ResNet-101 & ResNet-101 & 35.81 & -     & 36.39 & 26.77 & 14.69 &  5.41 &  0.65 \\
RefSegformer\cite{wu2024toward}\pub{TIP2024}      & Swin-B     & BERT       & 55.49 & 60.91 & 62.86 & 54.04 & 43.81 & 29.62 &  9.22 \\
LAVT\cite{yang2022lavt}\pub{CVPR2022}             & Swin-B     & BERT       & 56.46 & 61.48 & 63.46 & 54.89 & 45.31 & 31.78 & 12.03 \\
LGCE\cite{yuan2024rrsis}\pub{TGRS2024}            & Swin-B     & BERT       & 56.78 & 61.20 & 64.46 & 56.04 & 45.61 & 32.38 & 11.53 \\
RSRefSeg\cite{chen2025rsrefseg1}\pub{IGARSS2025}  & Swin-B     & BERT       & 50.02 & 53.16 & 54.59 & 46.72 & 37.84 & 26.82 & 14.34 \\
RMSIN\cite{liu2024rotated}\pub{CVPR2024}          & Swin-B     & BERT       & 60.10 & \uline{61.63} & \uline{68.67} & 60.75 & 51.53 & 37.04 & 12.93 \\
FIANet\cite{lei2024exploring}\pub{TGRS2024}       & Swin-B     & BERT       & \uline{61.32} & \textbf{62.97} & \textbf{70.13} & \textbf{62.16} & \uline{52.23} & 38.15 & 13.48 \\
\midrule

\rowcolor{gray!15}
\multicolumn{10}{l}{\textit{\textbf{LVLMs with Single-Temporal Pretraining, Without Fine-tuning}}} \\
LISA-7B\cite{lai2024lisa}\pub{CVPR2024}           & CLIP-L     & Vicuna-7B            &  4.88 &  5.11 &  1.78 &  0.79 &  0.40 &  0.20 &  0.00 \\
LISA-13B\cite{lai2024lisa}\pub{CVPR2024}          & CLIP-L     & Vicuna-13B           &  2.58 &  2.61 &  0.99 &  0.59 &  0.40 &  0.40 &  0.00 \\
GSVA-7B\cite{xia2024gsva}\pub{CVPR2024}           & CLIP-L     & Vicuna-7B            & 16.60 & 16.67 & 11.29 &  7.13 &  5.94 &  4.36 &  1.78 \\
GSVA-13B\cite{xia2024gsva}\pub{CVPR2024}          & CLIP-L     & Vicuna-13B           & 22.74 & 22.31 & 18.42 & 14.65 &  9.70 &  5.94 &  0.59 \\
GLAMM-7B\cite{rasheed2024glamm}\pub{CVPR2024}     & CLIP-L     & Vicuna-7B            & 10.07 &  7.70 &  7.92 &  5.15 &  2.97 &  1.19 &  0.99 \\
GeoPixel-7B\cite{shabbir2025geopixel}\pub{ICML2025} & CLIP-L  & InternLM-XComposer2.5 & 12.64 & 12.83 &  6.42 &  3.86 &  1.85 &  0.85 &  0.20 \\
UniGeoSeg\cite{ni2025unigeoseg}\pub{CVPR2026}     & Swin-B     & Phi-1.5              & 14.67 & 10.47 & 12.63 & 10.48 &  7.67 &  5.36 &  2.71 \\
SegEarth-R1\cite{li2025segearth}\pub{Arkiv2026}    & Swin-B     & Phi-1.5              &  5.77 &  4.96 &  2.81 &  2.01 &  1.10 &  0.80 &  0.30 \\

\midrule

\rowcolor{gray!15}
\multicolumn{10}{l}{\textit{\textbf{LVLMs with Single-Temporal Pretraining,  Fine-tuned on MTRefSeg-21K}}} \\
LISA-7B\cite{lai2024lisa}\pub{CVPR2024}           & CLIP-L     & Vicuna-7B            & 55.59 & 55.55 & 58.02 & 52.67 & 44.55 & 36.83 & 24.36 \\
LISA-13B\cite{lai2024lisa}\pub{CVPR2024}          & CLIP-L     & Vicuna-13B           & 58.10 & 56.13 & 60.00 & 55.45 & 49.31 & 40.79 & \uline{28.12} \\
GSVA-7B\cite{xia2024gsva}\pub{CVPR2024}           & CLIP-L     & Vicuna-7B            & \textbf{61.55} & 59.58 & 66.93 & 58.42 & 50.50 & \textbf{43.17} & 26.53 \\
GSVA-13B\cite{xia2024gsva}\pub{CVPR2024}          & CLIP-L     & Vicuna-13B           & 59.91 & 58.59 & 62.38 & 56.24 & 50.30 & \textbf{43.17} & \textbf{28.71} \\
GLAMM-7B\cite{rasheed2024glamm}\pub{CVPR2024}     & CLIP-L     & Vicuna-7B            & 60.08 & 58.10 & 64.16 & 57.43 & 50.30 & 40.40 & 25.94 \\
GeoPixel-7B\cite{shabbir2025geopixel}\pub{ICML2025} & CLIP-L   & InternLM-XComposer2.5 & 17.47 & 19.76 & 10.18 &  6.37 &  3.61 &  1.90 &  0.55 \\
SegEarth-R1\cite{li2025segearth}\pub{Arkiv2026}    & Swin-B     & Phi-1.5              & 54.95 & 54.86 & 60.37 & 54.11 & 47.44 & 37.07 & 19.04 \\
UniChange\cite{zhang2025unichange}\pub{CVPR2026}  & CLIP-L     & Vicuna-7B            & 41.51 & 39.63 & 41.39 & 35.84 & 27.72 & 19.60 & 10.89 \\
\rowcolor{yellow!20}
Ours                                    & Swin-B     & Phi-1.5              & 59.87 & 59.90 & 66.72 & \uline{61.15} & \textbf{53.93} & \uline{43.01} & 23.26 \\
\bottomrule
\end{tabular}
}
\end{table*}

%% file: table/0_MainTable_rs_0509.tex
\begin{table*}[t]
\centering
\scriptsize
\renewcommand{\arraystretch}{0.88}
\caption{
Comparison of different methods under the \textbf{RS} domain setting, where models are trained and evaluated within the remote-sensing domain.
This setting evaluates in-domain performance on remote-sensing change referring segmentation, where the training and validation data share similar aerial-view appearance, object scale, and spatial layout distributions.
}
\label{tab:rs_to_rs}
\resizebox{\textwidth}{!}{
\begin{tabular}{l l l c c c c c c c}
\toprule
\multirow{2}{*}{Model} & \multirow{2}{*}{Vision Encoder} & \multirow{2}{*}{Text Encoder} & \multicolumn{7}{c}{RS Train $\rightarrow$ RS Val} \\
\cmidrule(lr){4-10}
 &  &  & mIoU & cIoU/oIoU & Pr@50 & Pr@60 & Pr@70 & Pr@80 & Pr@90 \\
\midrule
\rowcolor{gray!15}
\multicolumn{10}{l}{\textit{\textbf{Specialist Methods, Trained on MTRefSeg-21K}}} \\
CRIS-50\cite{wang2022cris}\pub{CVPR2022}          & ResNet-50  & ResNet-50            & 55.34 & -     & 60.28 & 54.61 & 47.31 & 32.91 & 17.61 \\
CRIS-101\cite{wang2022cris}\pub{CVPR2022}         & ResNet-101  & ResNet-101            & 56.38 & -     & 60.78 & 54.81 & 47.65 & 34.19 & 18.15 \\
RefSegformer\cite{wu2024toward}\pub{TIP2024}      & Swin-B     & BERT                  & 64.54 & 67.02 & 70.33 & 65.61 & 59.07 & 49.56 & 25.70 \\
LAVT\cite{yang2022lavt}\pub{CVPR2022}             & Swin-B     & BERT                  & 66.57 & 68.61 & 72.48 & 66.89 & 60.22 & 51.19 & 27.55 \\
LGCE\cite{yuan2024rrsis}\pub{TGRS2024}            & Swin-B     & BERT                  & 66.79 & \textbf{68.57} & 72.55 & 67.74 & \uline{61.41} & \uline{52.09} & \uline{28.67} \\
RSRefSeg\cite{chen2025rsrefseg1}\pub{IGARSS2025}  & Swin-B     & BERT                  & 62.22 & 59.06 & 66.49 & 61.58 & 55.60 & 47.27 & 26.79 \\
RMSIN\cite{liu2024rotated}\pub{CVPR2024}          & Swin-B     & BERT                  & 66.45 & \uline{68.39} & 72.52 & 67.14 & 60.58 & 50.61 & 26.25 \\
FIANet\cite{lei2024exploring}\pub{TGRS2024}       & Swin-B     & BERT                  & \uline{67.06} & 68.38 & \uline{73.18} & \uline{68.02} & 61.27 & 50.98 & 26.21 \\

\midrule
\rowcolor{gray!15}
\multicolumn{10}{l}{\textit{\textbf{LVLMs with Single-Temporal Pretraining, Without Fine-tuning}}} \\
LISA-7B\cite{lai2024lisa}\pub{CVPR2024}           & CLIP-L     & Vicuna-7B             &  5.89 &  8.49 &  1.11 &  0.55 &  0.12 &  0.06 &  0.00 \\
LISA-13B\cite{lai2024lisa}\pub{CVPR2024}          & CLIP-L     & Vicuna-13B            &  2.35 &  2.51 &  0.68 &  0.49 &  0.49 &  0.43 &  0.43 \\
GSVA-7B\cite{xia2024gsva}\pub{CVPR2024}           & CLIP-L     & Vicuna-7B             & 12.93 & 13.95 &  5.29 &  3.13 &  1.60 &  0.80 &  0.25 \\
GSVA-13B\cite{xia2024gsva}\pub{CVPR2024}          & CLIP-L     & Vicuna-13B            & 14.32 & 15.06 &  7.50 &  4.06 &  2.03 &  1.04 &  0.12 \\
GLAMM-7B\cite{rasheed2024glamm}\pub{CVPR2024}     & CLIP-L     & Vicuna-7B             & 11.78 & 10.09 &  7.79 &  6.95 &  5.84 &  4.12 &  3.32 \\
GeoPixel-7B\cite{shabbir2025geopixel}\pub{ICML2025} & CLIP-L  & InternLM-XComposer2.5 & 10.24 & 12.15 &  5.16 &  3.54 &  2.31 &  1.65 &  1.06 \\
UniGeoSeg\cite{ni2025unigeoseg}\pub{CVPR2026}     & Swin-B     & Phi-1.5               & 14.06 &  6.87 & 12.32 & 11.11 & 10.14 &  9.09 &  7.83 \\
SegEarth-R1\cite{li2025segearth}\pub{Arkiv2026}    & Swin-B     & Phi-1.5               &  7.72 &  4.32 &  5.78 &  5.08 &  4.41 &  3.81 &  3.40 \\

\midrule

\rowcolor{gray!15}
\multicolumn{10}{l}{\textit{\textbf{LVLMs with Single-Temporal Pretraining,  Fine-tuned on MTRefSeg-21K}}} \\
LISA-7B\cite{lai2024lisa}\pub{CVPR2024}           & CLIP-L     & Vicuna-7B             & 50.19 & 47.22 & 52.92 & 43.39 & 33.56 & 24.71 & 17.70 \\
LISA-13B\cite{lai2024lisa}\pub{CVPR2024}          & CLIP-L     & Vicuna-13B            & 51.18 & 45.24 & 54.95 & 44.74 & 33.68 & 25.63 & 18.81 \\
GSVA-7B\cite{xia2024gsva}\pub{CVPR2024}           & CLIP-L     & Vicuna-7B             & 53.76 & 47.66 & 58.02 & 48.06 & 37.74 & 28.15 & 20.34 \\
GSVA-13B\cite{xia2024gsva}\pub{CVPR2024}          & CLIP-L     & Vicuna-13B            & 52.41 & 47.92 & 56.98 & 45.79 & 35.34 & 26.86 & 17.89 \\
GLAMM-7B\cite{rasheed2024glamm}\pub{CVPR2024}     & CLIP-L     & Vicuna-7B             & 54.36 & 47.25 & 58.51 & 48.19 & 37.25 & 27.90 & 19.67 \\
GeoPixel-7B\cite{shabbir2025geopixel}\pub{ICML2025} & CLIP-L  & InternLM-XComposer2.5 & 17.66 & 15.19 & 11.62 &  9.75 &  8.30 &  7.01 &  6.32 \\
SegEarth-R1\cite{li2025segearth}\pub{Arkiv2026}    & Swin-B     & Phi-1.5               & 62.38 & 62.57 & 67.59 & 62.72 & 56.64 & 46.00 & 25.72 \\
UniChange\cite{zhang2025unichange}\pub{CVPR2026}  & CLIP-L     & Vicuna-7B             & 57.80 & 46.84 & 62.38 & 55.75 & 47.70 & 38.05 & 20.22 \\
\rowcolor{yellow!20}
Ours                                    & Swin-B     & Phi-1.5              & \textbf{68.92} & 67.79 & \textbf{74.32} & \textbf{69.82} & \textbf{64.66} & \textbf{56.24} & \textbf{32.95} \\
\bottomrule
\end{tabular}
}
\end{table*}

%% file: sec/5_conclusion.tex
\section{Conclusion}
\label{sec:conclusion}
In this paper, we introduced Multi-temporal Referring Segmentation (MTRS), a new task that requires models to segment language-described temporal changes from multi-temporal images. To support this task, we proposed CRAFT-Agent and constructed MTRefSeg-21K, a benchmark containing high-quality bi-image--text--mask annotations across normal-scene and remote-sensing domains. We further benchmarked representative VLMs and LVLMs, showing that existing single-temporal vision-language models struggle with temporal correspondence reasoning and language-guided change localization. To address these challenges, we proposed MTRefSeg-R1, a change-aware LVLM framework with vision-only temporal-change pretraining and full multimodal fine-tuning. Extensive experiments demonstrate that MTRefSeg-R1 achieves superior performance over existing baselines. We hope this work will facilitate future research on language-guided multi-temporal visual understanding.

\textbf{Limitations.}
Although MTRefSeg-R1 improves multi-temporal referring segmentation, it still relies mainly on bi-temporal inputs and may struggle with severe viewpoint changes, extremely small changed objects, and ambiguous referring expressions. Future work will extend MTRS to longer temporal sequences, improve robustness under large cross-view variations, and explore more efficient training strategies for large-scale deployment.
\clearpage